\documentclass{article}

\usepackage[preprint]{corl_2024} 

\usepackage[numbers]{natbib}
\usepackage{multicol}
\usepackage{hyperref}
\usepackage{amsmath, amsfonts, amsthm, amssymb}
\usepackage{mathabx}
\usepackage{algorithm}
\usepackage[noend]{algpseudocode}
\usepackage{glossaries}
\usepackage{graphicx}
\usepackage{bm}

\newcommand{\revise}[1]{\textcolor{black}{#1}}
\DeclareMathOperator*{\argmax}{arg\,max}

\DeclareMathOperator{\diag}{diag}

\newcommand\scalemath[2]{\scalebox{#1}{\mbox{\ensuremath{\displaystyle #2}}}}

\DeclareMathOperator*{\xpt}{\mathbb{E}}

\newcommand{\expect}[2]{\xpt_{\substack{#1}} \left[ #2 \right]}

\newcommand{\state}{s}
\newcommand{\act}{a}
\newcommand{\searchdist}{\pi}

\newcommand{\q}{\bm{q}}
\newcommand{\dq}{\dot{\q}}
\newcommand{\ddq}{\ddot{\q}}

\newcommand{\traj}{\bm{\tau}}

\newcommand{\constraint}{c}
\newcommand{\Constraint}{\bm{c}}

\newcommand{\bLMi}{\bar{C}_i}

\newcommand{\tasks}{\mathcal{T}}
\newcommand{\task}{T}
\newcommand{\params}{\theta}
\newcommand{\func}{f}

\newcommand{\BasisFunctions}{\bm{\Phi}}
\newcommand{\cps}{\bm{w}}

\newcommand{\reals}{\mathbf{R}}
\newcommand{\nq}{n_{q}}
\newcommand{\nb}{n_{b}}
\newcommand{\ns}{n_{s}}
\newcommand{\knots}{\bm{k}}
\newcommand{\knot}{k}

\newcommand{\Jtask}{J_{\text{task}}}
\newcommand{\Ltask}{L_{\text{task}}}

\DeclareMathOperator{\clip}{clip}

\DeclareMathOperator{\selu}{SELU}

\newcommand{\scale}{a}

\newacronym{rl}{RL}{Reinforcement Learning}
\newacronym{drl}{DRL}{Deep Reinforcement Learning}
\newacronym[plural=MPs, firstplural=Motion primitives (MPs)]{mp}{MP}{Motion Primitive}
\newacronym{hrl}{HRL}{Hierarchical Reinforcement Learning}
\newacronym{saferl}{SafeRL}{Safe Reinforcement Learning}
\newacronym{avi}{AVI}{Approximate Value-Iteration}
\newacronym{api}{API}{Approximate Policy-Iteration}
\newacronym[plural=MDPs, firstplural=Markov Decision Processes (MDPs)]{mdp}{MDP}{Markov Decision Process}
\newacronym{cmdp}{CMDP}{Constrained Markov Decision Processes}
\newacronym{safeexp}{SafeExp}{Safe Exploration}
\newacronym{kl}{KL}{Kullback-Leibler Divergence}
\newacronym{gae}{GAE}{Generalized Advantage Estimation}
\newacronym{papi}{PAPI}{Projections for Approximate Policy Iteration}
\newacronym{her}{HER}{Hindsight Experience Replay}
\newacronym{ham}{HAM}{Hierarchy of Abstract Machines}
\newacronym{mom}{MOM}{Measure of Manipulability}
\newacronym{bo}{BO}{Bayesian Optimization}
\newacronym{hebo}{HEBO}{Heteroscedastic Evolutionary Bayesian Optimisation}
\newacronym{ucb}{UCB}{Upper Confidence Bound}
\newacronym{pi}{PI}{Probability of Improvement}
\newacronym{ei}{EI}{Expected Improvement}
\newacronym{nl}{NLP}{Nonlinear Programming}
\newacronym{lp}{LP}{Linear Programming}
\newacronym{qp}{QP}{Quadratic Programming}
\newacronym{aqp}{AQP}{Anchored Quadratic Programming}
\newacronym{ode}{ODE}{Ordinary Differential Equation}
\newacronym{atacom}{ATACOM}{Acting on the TAngent Space of the COnstraint Manifold}
\newacronym{ppolag}{PPO-Lag}{PPO-Lagrangian}
\newacronym{trpolag}{TRPO-Lag}{TRPO-Lagrangian}
\newacronym{pcpo}{PCPO}{Projection-Based Constrained Policy Optimization}
\newacronym{ivp}{IVP}{Initial Value Problem}
\newacronym{rref}{RREF}{Reduced Row Echlon Form}
\newacronym{rcef}{RCEF}{Reduced Column Echlon Form}
\newacronym{cpo}{CPO}{Constrained Policy Optimization}
\newacronym{trpo}{TRPO}{Trust Region Policy Optimization}
\newacronym{rmp}{RMP}{Riemannian Motion Policies}
\newacronym{dnn}{DNN}{Deep Neural Networks}
\newacronym{sdf}{SDF}{Signed Distance Function}
\newacronym{redsdf}{ReDSDF}{Regularized Deep Signed Distance Fields}
\newacronym{apf}{APF}{Artificial Potential Fields}
\newacronym{hri}{HRI}{Human-Robot Interaction}
\newacronym{poi}{PoI}{Point of Interest}
\newacronym{pdf}{PDF}{Probability Density Function}
\newacronym{cdf}{CDF}{Cumulative Distribution Function}

\newacronym{mpc}{MPC}{Model Predictive Control}
\newacronym{dcs}{DCS}{Directly Controllable State}
\newacronym{dus}{DUS}{Directly Uncontrollable State}

\newacronym{cbf}{CBF}{Control Barrier Function}

\newacronym{var}{VaR}{Value-at-Risk}
\newacronym{cvar}{CVaR}{Conditional Value-at-Risk}

\newacronym{sac}{SAC}{Soft Actor Critic}
\newacronym{ppo}{PPO}{Proximal Policy Optimization}

\newacronym{promp}{ProMP}{Probabilistic Movement Primitives}
\newacronym{prodmp}{ProDMP}{Probabilistic Dynamic Movement Primitives}
\newacronym{dmp}{DMP}{Dynamic Movement Primitives}
\newacronym{rtp}{RTP}{Residual Trajectory Primitives}
\newacronym{cnpb}{CNP-B}{Constrained Neural motion Planning with B-splines}
\newacronym{ours}{CNP3O}{Constrained Neural motion Planning with PPO}

\newacronym{fc}{FC}{fully connected}

\title{Bridging the gap between Learning-to-plan, Motion Primitives and Safe Reinforcement Learning}

\author{
  Piotr Kicki$^{1,2}$, Davide Tateo$^3$, Puze Liu$^3$, Jonas Guenster$^3$, Jan Peters$^3$, Krzysztof Walas$^{1,2}$\\
  $^1$IDEAS NCBR, Warsaw, Poland\\
  $^2$Institute of Robotics and Machine Intelligence, Poznan University of Technology, Poland\\
  $^3$Department of Computer Science, Technische Universitat Darmstadt, Germany\\
  \texttt{piotr.kicki@ideas-ncbr.pl} \\
}

\begin{document}
\maketitle


\begin{abstract}
    Trajectory planning under kinodynamic constraints is fundamental for advanced robotics applications that require dexterous, reactive, and rapid skills in complex environments. These constraints, which may represent task, safety, or actuator limitations, are essential for ensuring the proper functioning of robotic platforms and preventing unexpected behaviors. Recent advances in kinodynamic planning demonstrate that learning-to-plan techniques can generate complex and reactive motions under intricate constraints. However, these techniques necessitate the analytical modeling of both the robot and the entire task, a limiting assumption when systems are extremely complex or when constructing accurate task models is prohibitive.
    This paper addresses this limitation by combining learning-to-plan methods with reinforcement learning, resulting in a novel integration of black-box learning of motion primitives and optimization. We evaluate our approach against state-of-the-art safe reinforcement learning methods, showing that our technique, particularly when exploiting task structure, outperforms baseline methods in challenging scenarios such as planning to hit in robot air hockey. This work demonstrates the potential of our integrated approach to enhance the performance and safety of robots operating under complex kinodynamic constraints.    
\end{abstract}



\section{Introduction}
Nowadays, robots are capable of complex dynamic tasks such as table tennis~\cite{mulling2011biomimetic,buchler2022learning}, juggling~\citep{ploeger2021high,ploeger2022controlling} or diabolo~\citep{von2021analytical}, and play sports such as tennis~\citep{zaidi2023athletic} or soccer~\citep{haarnoja2024learning}. Current planning and learning methods are sufficient for most of these tasks, as the robot's movement is relatively free in the workspace, and they are not required to comply with stringent tasks, hardware, and safety constraints. However, these requirements become fundamental if we want to deal with real robotics tasks in unstructured environments in the real world. Therefore, the lack of competitive techniques to efficiently plan trajectories in unknown and unstructured environments under constraints strongly limits the applicability of modern robotics and learning frameworks to tasks beyond the lab setting.

To fix this gap, \citet{altman1998constrained} introduced the \gls{cmdp} framework, and, based on this setting, researchers in machine learning developed the \gls{saferl} techniques to efficiently solve the \gls{cmdp} problem without full knowledge of the environment. 
However, these methods are not able to scale effectively to complex tasks. Furthermore, since most of these approaches learn black-box approximations, they do not allow effective exploitation of domain knowledge. 

\revise{
\glspl{mp} are a technique that enables efficient encoding of domain knowledge in the policy not fully exploited in the \gls{saferl} context. Interestingly, these approaches plan full trajectories, allowing the agent to check for the safety of the whole trajectory before executing it, preventing the agent from reaching states that may heavily violate the constraints. However, given that \glspl{mp} allows encoding the safety features in the trajectory, the literature lacks general approaches to deal with a general set of constraints. Unfortunately, this forces the user to rely heavily on hand-crafting \glspl{mp}, possibly resulting in suboptimal solutions.}


A valid alternative to the methods above is to exploit learning-to-plan methods that can generate full trajectories, as in the \gls{mp} setting, while imposing constraints during the planning time or even in the learning process. However, learning-to-plan approaches require the full knowledge of the task being optimized. \revise{This assumption is very different from the \gls{saferl} setting, where we assume to have access only to environment rollouts.}

\revise{In this paper, we draw connections between these aforementioned fields and we show how to extend the learning-to-plan methods to exploit the knowledge of the constraints without requiring full knowledge of the environment.} We extend the framework presented in~\cite{kicki2023fast} to the \gls{rl} setting, resulting in a hybrid method that shares many common ideas with the \gls{saferl}, \gls{mp}, and learning-to-plan approaches. In particular, the contributions of this paper are i. a novel algorithm to learn how to generate \gls{mp}-based trajectories under known constraints; ii. an analysis of different \glspl{mp}, where we show that the B-splines are particularly useful for learning under constraints; iii. practical guidelines on how to properly impose domain knowledge on \glspl{mp} in the learning-to-plan setting.

We evaluate our approach in two challenging tasks, i.e., moving a heavy vertically-oriented object with a manipulator and a robotic air hockey hitting task. The first task is challenging as it pushes the limits of robot actuators, while the second task requires learning a highly dynamic motion under complicated constraints and sensitive objective functions. 
Finally, we show that we could deploy the proposed method in a real robot air hockey setup.

\subsection*{Related work}
The literature on safety is quite broad, and there are many different \gls{saferl} approaches that tackle the \gls{cmdp} problem~\citep{altman1998constrained,altman1999constrained}. 
The first and most popular technique to solve this problem is the lagrangian relaxation approach\revise{~\cite{altman1998constrained, tessler2019reward, stooke2020responsive, ioannis2022guided, ding2021provably, borkar2014risk, ying2022towards, yang2023safety},} where the original task objective is mixed trough a lagrangian multiplier with the constraint cost.
Other alternative solutions instead rely on different ideas from the optimization literature, such as state augmentation~\cite{sootla2022enhancing}, the trust region methods~\cite{achiam2017constrained, kim2022efficient} and the interior point approach~\cite{liu2020ipo}. 
An alternative formulation of the problem is based on more control-theoretic insights. These approaches are based either on Lyapunov functions~\cite{chow2018lyapunov, chow2019lyapunov, sikchi2021lyapunov}, Control barrier functions~\cite{ames2019control,xiao2022high_order, taylor2020learning, cheng2019end} or reachability analysis~\cite{fisac2018general,shao2021reachability,selim2022safe,zheng2024safe}. The last category of solutions to safety problems is shielding techniques that correct potentially dangerous actions to be applied in the system~\cite{hans2008safe, garcia2012safe,berkenkamp2017safe,fisac2019bridging, pham2018optlayer,dalal2018safe, ames2019control,liu2022robot, emam2022safe,liu2023safe}. Some approaches in these two last categories can guarantee safety at every step of the learning process but require prior knowledge of system dynamics, backup policies, or unsafe interaction datasets.

A very classical approach to introduce safety into robot actions is motion planning~\cite{lavalle2006planning}. However, typically classical planning methods struggle to meet the real-time requirements for complex tasks. Thus, we observe the growing interest in the motion planning community for exploiting learning to improve and speed-up planning~\cite{ qureshi2019mpnet, kicki2022speedingup, kicki2023fast, johnson2023transformers, joao2023diffusion}. Many approaches that apply learning to motion planning utilize learning to bias the prediction of the next segment of the solution~\cite{qureshi2019mpnet, kicki2021eaai, johnson2023transformers}. However, methods of this type struggle to generate feasible dynamic trajectories for complex problems with challenging constraints~\cite{kicki2023fast}. An interesting alternative to combine planning and learning is so-called \textit{planning-as-inference}, an example of which may be the use of diffusion models to optimize whole trajectories~\cite{joao2023diffusion} or inferring the whole motion planning problem solutions with a neural network~\cite{kicki2022speedingup, osa2022mp}. In this spirit, authors of~\cite{kicki2023fast} introduced a machine learning-based method for planning trajectories that satisfy a variety of constraints to ensure the system safety. However, the approach introduced in~\cite{kicki2023fast} requires the differentiability of the task loss function, which is not the case for many interesting real-world problems.

A very important aspect of learning how to generate trajectories is their representation. While it is possible to follow the auto-regressive approach~\cite{qureshi2019mpnet, kicki2021eaai}, methods that utilize more structured trajectory representations seem to offer more benefits, like compactness, reduced planning time and boundary conditions satisfaction~\cite{kicki2023fast, osa2022mp, kicki2022speedingup, promp, li2023prodmp}. One of the most common approaches for trajectory representation in learning-based solutions are \glspl{mp}~\cite{promp, dmp, li2023prodmp, osa2022mp, lee2023equivariant}, as they offer a very flexible and compact representation. To the most widespread \glspl{mp} belongs \gls{promp}~\cite{promp} and \gls{dmp}~\cite{dmp}, which benefits were recently combined in \gls{prodmp}~\cite{li2023prodmp}, allowing for some simple boundary constraints satisfaction and efficient computation, which is also possible with the use of \glspl{mp} presented in~\cite{osa2022mp}. However, by far the most composite and flexible solution was introduced in~\cite{kicki2023fast}, where B-splines were used to construct a trajectory. This representation allows for imposing boundary conditions on the trajectory and its derivatives and introduces great flexibility in terms of the trajectory timing.

This paper, builds upon the approach introduced in~\cite{kicki2023fast}. We show that trajectory representation from~\cite{kicki2023fast} is in fact a \gls{mp}, and we extend the learning-to-plan method that uses them to \gls{rl} setting via a hybrid approach that bridges the gap between learning-to-plan, \glspl{mp} and \gls{saferl}.





\section{Constrained Reinforcement Learning with Motion Primitives}
\vspace{-0.1em}
\label{sec:constrained_rl}
Our problem is to find a sampling distribution $\searchdist$ over trajectories $\traj$ maximizing the expected cumulative reward while satisfying the safety constraints $g, h$ over the whole trajectory. The performance objective is defined on a distribution of tasks $\tasks$ parameterized with a task definition $\task$. The resulting optimization problem is
\begin{align}
    \scalemath{0.9}{\argmax_{\searchdist} \quad\expect{\task \sim \tasks}{\expect{\traj \sim \searchdist}{\sum_{t=0}^{t_{\traj}} \gamma^t r_\task(\state_t,\act(\traj(t)))}}} & 
    & \scalemath{0.9}{\text{s.t.}  \qquad}  &
    \begin{aligned}
        \scalemath{0.9}{g_i(\traj(t), t) = 0} & \;\scalemath{0.9}{\forall t, \forall i\in\lbrace 1,\dots, N \rbrace},\\
         \scalemath{0.9}{h_j(\traj(t), t) \leq 0} 
         & \;\scalemath{0.9}{\forall t, \forall j\in\lbrace 1,\dots, M \rbrace},
    \end{aligned}
    \label{eq:general_problem_formulation}
\end{align}
where $\sum_{t=0}^{t_{\traj}} \gamma^t r_\task(\state_t,\act(\traj(t)))$ is the discounted task reward $\Jtask$. 
We assume that the action $\act$ is computed by a controller based on the trajectory representation $\traj(t)$ and the current robot state.

\begin{figure}[t]
    \centering
    \includegraphics[width=0.98\linewidth]{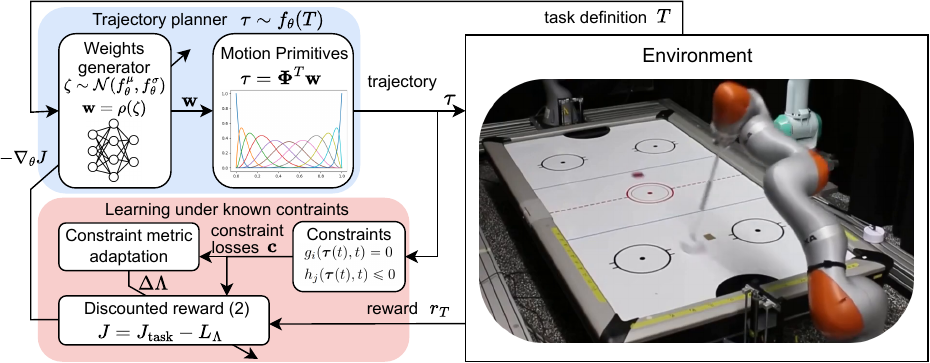}
    \vspace{-2mm}
    \caption{Overview of the proposed constrained trajectory generation method.}
    \label{fig:main_scheme}
    \vspace{-1.3em}
\end{figure}

To solve this problem, we propose to generate search distribution $\searchdist = \func_\params(\task)$ using a function $\func_\params$, parametrized with $\params$, based on the given task definition $\task$. We formalize this function as a linear combination of \glspl{mp} with weights computed with a nonlinear transformation $\rho$ of the samples drawn from the normal distribution, in which mean and standard deviation are determined by a neural network. Generated trajectories are evaluated in the environment, and based on the trajectory and the reward from the environment, we optimize the neural network weights $\params$ to maximize the task reward and minimize the constraints violations. The overview of the proposed solution is presented in Figure~\ref{fig:main_scheme}, while we discuss its important components in the next sections.

\subsection{Learning under known constraints}
This section, describes the proposed solution for solving the constrained optimization problem defined in~\eqref{eq:general_problem_formulation}. One of the most common approaches to address this challenge is the method of Lagrange multipliers~\cite{altman1998constrained, ying2022towards, yang2023safety}. Inspired by this approach, we propose to relax the constraints, assuming some acceptable violation budget, and include them in the objective with learnable scaling factors. In this paper, the optimization with the extended objective is interleaved with the adaptation of the Lagrange multipliers associated with specific constraints, which can be interpreted as learning the metric of the constraint manifold. We exploit the knowledge about the considered system constraints and the fact that typical robot constraints are differentiable w.r.t. trajectory $\traj$ to optimize the neural network to satisfy them using their analytical gradient.
Moreover, we show that it is possible to include a range of different constraints without aggregating them into a single constraint cost function, which enables better handling of multiple constraints. In the following, we present the \gls{ours} algorithm \revise{(see Algorithm~\ref{alg:CNP3O})}, and we show how it can be derived by extending the \gls{cnpb} approach~\cite{kicki2023fast} to the \gls{rl} setting.


First, following~\cite{kicki2023fast}, we transform both equality $g_i(\traj(t), t) = 0$ and inequality constraints $h_j(\traj(t), t) \leq 0$ into inequality constraints of the form $\constraint_i(\traj(t), t) \leq \bLMi$, where $\constraint_i(\traj(t), t)$ represents the $i$-th constraint violation of the trajectory $\traj(t)$ and $\bLMi$ is the assumed $i$-th constraint violation budget.
Then, we need to incorporate these constraints into the task objective. 
To do so, we rewrite the objective function from \eqref{eq:general_problem_formulation} into
\begin{equation}
    \scalemath{0.9}{J = \expect{\task \sim \tasks}{\expect{\revise{\zeta} \sim \func_\params(\task)}{\sum_{t=0}^{t_{\traj}} \gamma^t r_\task(\state_t,\act(\revise{\rho(\zeta(t)))})} - \Constraint^T(\rho(\func_\params^\mu(\task)), t)\Lambda\Constraint(\rho(\func_\params^\mu(\task)), t)}},
    \label{eq:J}
\end{equation}
where $\func_\params^\mu(\task)$ is the true mean of the distribution induced by the neural network $\func$ for a given task $\task$, $\rho$ is a nonlinear transformation of the samples $\zeta$ (for more details see Appendix~\ref{app:samples_transformation}), $\Lambda = \diag(\lambda_1, \lambda_2, \ldots, \lambda_{N+M})$ is the diagonal matrix of the Lagrange multipliers and $\Constraint^T(\revise{\rho(\func_\params^\mu(\task))}, t)\Lambda\Constraint(\revise{\rho(\func_\params^\mu(\task)}), t)$ is the manifold loss $L_\Lambda$.
In this way, we decouple learning how to solve a task from learning how to satisfy the constraints and allow one to optimize the constraint satisfaction against the neural network weights $\params$ directly, without the need for sampling. By doing so, we are not penalizing the constraints violation done by the samples of the trajectory distribution but only its mean. Therefore, to ensure safety in the real robot deployment, we assume that the mean trajectory will be used, while for learning on the real robot, one may not execute trajectories that violate the constraints too much.
We explicitly allow for imposing multiple constraints in a decoupled way, such that each of them has its own scaling factor $\lambda_{i}$. To ensure the positiveness of these scaling factors, we parameterize each of them with $\eta_i$ using $\lambda_i = \exp(\eta_i)$. These scaling parameters $\eta_i$ are updated based on the mean constraints violations observed in the previous set of episodes, which can be defined by $
    \Delta\eta_i = \alpha \log\left(\frac{\constraint_i + \beta \bLMi}{\bLMi}\right)
$, where $\alpha > 0$ is the constraint learning rate and $0 < \beta < 1$ bounds the rate of the $\eta_i$ decline.

The remaining part of the discounted reward $J$ is related to solving the task. To evaluate the sample task reward, we first generate the normal distribution $\mathcal{N}(\func_\theta^\mu(\task), \func_\sigma^\mu(\task))$ using neural network $\func_\params$, which predicts its mean $\mu$ and standard deviation $\sigma$ for a given task $\task$. Then, we sample from $\zeta\sim\mathcal{N}(\func_\theta^\mu(\task), \func_\sigma^\mu(\task))$, \revise{process these samples with the samples transformation function $\rho(\zeta)$,} and evaluate generated trajectories $\traj$ in the considered environment. Finally, we use the obtained accumulated rewards from the simulated episodes to optimize the neural network weights $\params$, using an episodic version of the \gls{ppo} algorithm due to its simplicity.

\begin{algorithm}[b]
    \caption{\gls{ours}}\label{alg:CNP3O}
    \footnotesize
    \begin{algorithmic}[1]
    \For{$k \gets 1$ to $N_{\text{epochs}}$}
        \For{$i \gets 1$ to $N_{\text{episodes}}$}
            \color{black}
            \State Sample task $\task$ from distribution $\tasks$
            \State Sample $\zeta$ from the distribution $\mathcal{N}(\func^\mu_\params(\task), \func^\sigma_\params(\task))$ and compute the sample trajectory $\traj = \BasisFunctions \rho(\zeta)$
            \State Evaluate $\traj$ in the environment to get $\Jtask$.
            \State Compute the value function $V_\psi(\task)$ of task $\task$
            \State Store $(\Jtask, \func^\mu_\params(\task), V_\psi(\task))$
        \EndFor
        \For{$i \gets 1$ to $N_{\text{fits}}$}
            \For{$j \gets 1$ to $N_{\text{batches}}$}
            \color{black}
                \State Sample a batch from the stored data
                \State Compute task loss $\Ltask$ using episodic PPO algorithm based on the $\Jtask$ and $V(\task)$
                \State Compute manifold loss $L_\Lambda$ based on the $\rho(\func^\mu_\params(\task))$
                \State Update the policy network weights $\params$ based on the gradient of $\Ltask + L_\Lambda$
                \State Update the value network weights $\psi$ based on the gradient of $(\Jtask - V_\psi(\task))^2$
            \EndFor
            \State Update manifold metric $\Lambda$
        \EndFor
    \EndFor
    \end{algorithmic}
\end{algorithm}



\subsection{Motion Primitives for Safe Reinforcement Learning}

One of the most important design decisions in the case of learning how to generate plans is the representation of the trajectories that the planner will generate. \glspl{mp}~\cite{osa2022mp, li2023prodmp} are very popular, general, and flexible trajectory representations that can be used for planning. They can be, in general, defined by $  \scalemath{0.9}{\q(s) = \BasisFunctions(s) \cps}$,
where the robot configuration $\q(s) \in \reals^{\nq}$, for given value of the phase variable $s$, is computed as a product of basis functions $\BasisFunctions(s) \in \reals^{\nq \times \nb}$, evaluated at $s$, and the weights vector $\cps \in \reals^{\nb}$. 
This formulation may describe diverse \glspl{mp}, such as \gls{promp}~\cite{promp}, \gls{prodmp}~\cite{li2023prodmp}, \gls{rtp}~\cite{osa2022mp}, just by defining basis functions $\BasisFunctions$ differently or adding some biases to the weights $\cps$. To interpret these \glspl{mp} as trajectories, we need to transform the dependency on the phase variable $s$ into dependency on time $t$. In the literature, it is typical to do it directly by assuming that $s=t$, or by a linear scaling, i.e., $s = \frac{t}{T_s}$, where $T_s > 0$ is the time scaling factor. 
Each of the aforementioned \glspl{mp} offers different useful properties from the motion planning point of view, such as the guarantee of connecting the initial $\q_0$ and target $\q_d$ configurations. However, we argue that much more may be offered by using the B-spline-based \glspl{mp}. In this work, we show that the trajectory representation proposed in~\cite{kicki2023fast} can be viewed as \gls{mp} and highlight its benefits over the existing \glspl{mp}-based approaches.

First, let us note that any spline function of order $n$ can be represented as a linear combination of $n$-th order B-splines, i.e. $\q(s) = \sum_i \cps_i B_{i, n}^{\knots}(s)$, where $B_{i, n}^{\knots}$ is the $n$-th order B-spline basis function defined between $\knot_i$ and $\knot_{i+n+1}$, where $\knots$ is a vector of knots that define domains of the B-spline basis functions. However, in general, the vector of knots $\knots$ may be an arbitrary non-decreasing sequence that partitions the domain of the represented function, such that its changes affect the shape of the \glspl{mp}, which would have to be then recomputed every time. 
To avoid this, we propose fixing the knots vector and limiting it to the range of $s \in [0; 1]$. 
Thanks to this, we can drop the dependency on knot vector $\knots$ and describe B-splines of given order $n$ as \glspl{mp} $q(s) = \BasisFunctions(s) \cps$, where $\BasisFunctions(s) = \begin{bmatrix} B_1(s) & B_2(s) & \cdots & B_n(s)\end{bmatrix}^T \cps$. Moreover, it is possible to precompute all of the basis functions in advance for a range of the phase variable values $\{0, \frac{1}{\ns-1}, \frac{2}{\ns - 1}, \ldots, 1\} \in \reals^{\ns}$ and create a tensor $\BasisFunctions \in \reals^{\ns \times \nq \times \nb}$, which allows for computing the movement as a matrix-vector product, similarly like it can be done for \gls{promp}~\cite{promp} and \gls{prodmp}~\cite{li2023prodmp}. 
While the choice of the knot vector $\knots$ may be arbitrary, we propose making subsequent knots equidistant to distribute the basis functions across the domain equally. Moreover, by fixing the first and last $n+1$ knots to be equal to 0 and 1, respectively, we can guarantee that the generated path will start in the point defined by $\cps_1$ and end in $\cps_{\nb}$.

Finally, we must transform the phase variable $s$ into the time $t$ to generate a trajectory.
Therefore, we introduce another B-spline function 
$r(s) = \left(\frac{dt}{ds}\right)^{-1}(s)$,
as done in~\cite{kicki2023fast}. This flexible time scaling allows us to flexibly control the derivatives of the resultant trajectory w.r.t. time and the overall trajectory duration, which is not so straightforward for the methods proposed in~\cite{promp, li2023prodmp, osa2022mp}. Moreover, this time parameterization allows one to directly impose the boundary constraints on the initial and target velocities, accelerations, and higher order derivatives up to the $d$-th derivative by analytically setting the $d+1$ boundary weights of the configuration. To the best of our knowledge, this flexibility in imposing boundary conditions is not present in any existing \gls{mp} framework.

Imposing boundary conditions is an important feature of the trajectory representations as it allows for connecting the subsequent trajectories, which is particularly important to concatenate trajectory segments, enable online replanning, and may be useful in incorporating some prior knowledge into the designed solution. An example of this feature is a task that features reaching a known goal. Then, we can easily impose this goal configuration as a prior for the solutions returned by the planner. Thanks to the ability to impose boundary conditions also on the derivatives, we can incorporate strong priors that enable ending the motion with some predefined velocity, useful in tasks that require hitting, moving, or tossing objects. 
This is also important for connecting trajectories smoothly, e.g., making $\dddot{q}(t)$ continuous we can ensure the smoothness of controls, despite the change of the trajectory segment.
Thanks to this capability, we can generate both single trajectories and smoothly connected sequences of trajectories passing through the via-points. Further discussion on imposing prior knowledge on \glspl{mp}-based trajectories is provided in Appendix~\ref{app:prior_knowledge}.


\section{Experimental Evaluation}
\label{sec:results}
In this section, we evaluate the performance of our proposed method in the same two tasks defined in~\cite{kicki2023fast}, \revise{using 10 seeds}. In both environments, the objective is to control a Kuka Iiwa 14 manipulator.
Our experimental evaluation has two objectives. 
First, we want to evaluate the benefit of our framework against classical \gls{saferl} approaches. \revise{For this reason, we consider following important baselines in the area, \gls{atacom}~\cite{liu2022robot,liu2023safe,liu2024safe} -- a model-based approach, model-free methods \gls{ppolag} and \gls{trpolag}~\cite{achiam2019benchmarking}, and a projection based extension of \gls{trpolag} -- \gls{pcpo}~\cite{tsung2020pcpo}. All of these approaches learn a classical neural policy to control the robot. However, \gls{atacom} ensures safety by exploiting the model of the robot and the constraints, \gls{ppolag} and \gls{trpolag} implement a model-free lagrangian optimization technique, while \gls{pcpo} additionally projects the policy on the constraint set.}
The second objective of the conducted experiments is to evaluate which \gls{mp} is more adequate for the learning-to-plan setting. Therefore, we compare the B-splines~\cite{kicki2023fast}, \gls{promp}~\cite{promp} and \gls{prodmp}~\cite{li2023prodmp}.
\begin{figure}[b!]
    \vspace{-0.5cm}
    \centering
    \setlength{\tabcolsep}{1.5pt}
    \begin{tabular}{ccc}
         \scriptsize \textbf{a) Heavy Object w/o Prior Knowledge} & \scriptsize \textbf{b) Heavy Object w/ Prior Knowledge} &
         \scriptsize \textbf{c) Air Hockey Hitting}\\
         \includegraphics[width=0.32\textwidth]{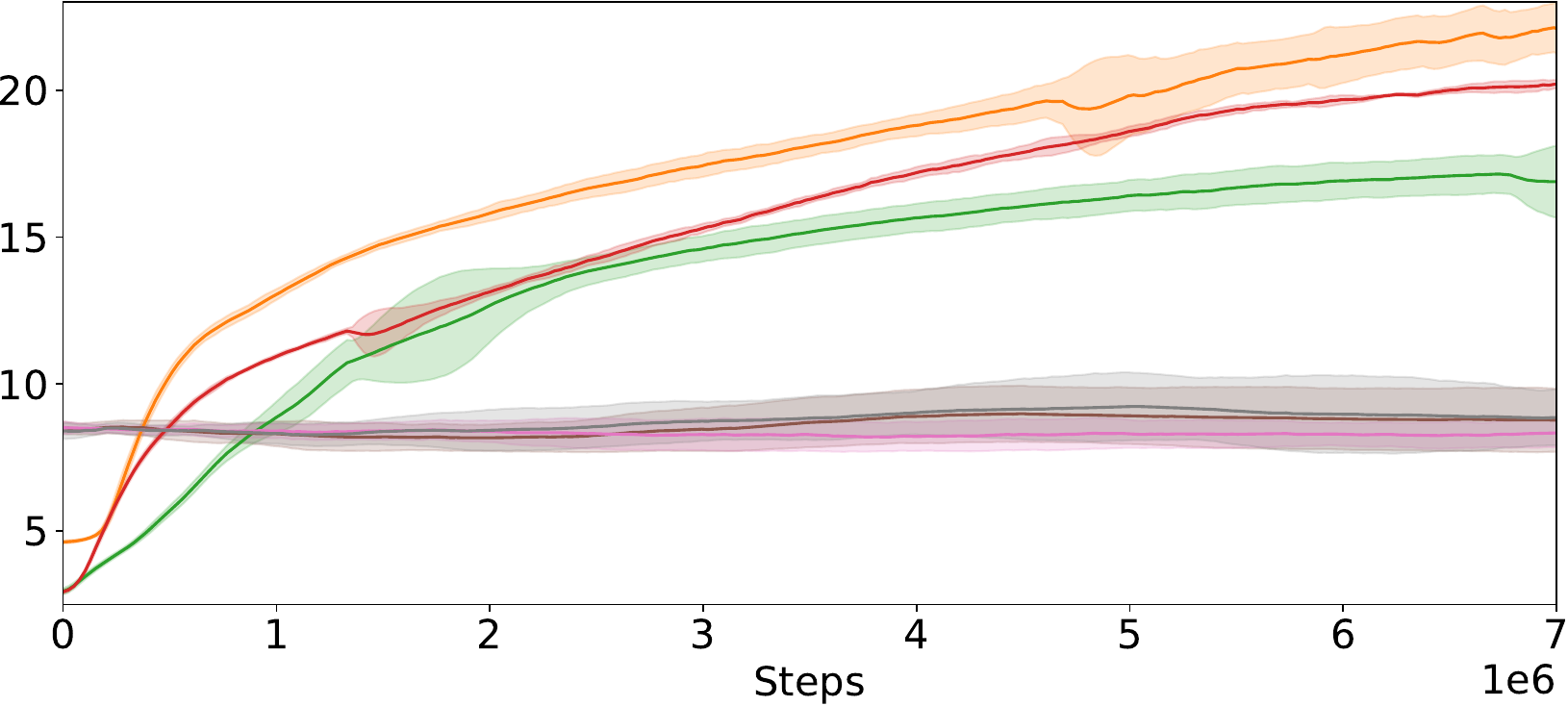} & \includegraphics[width=0.32\textwidth]{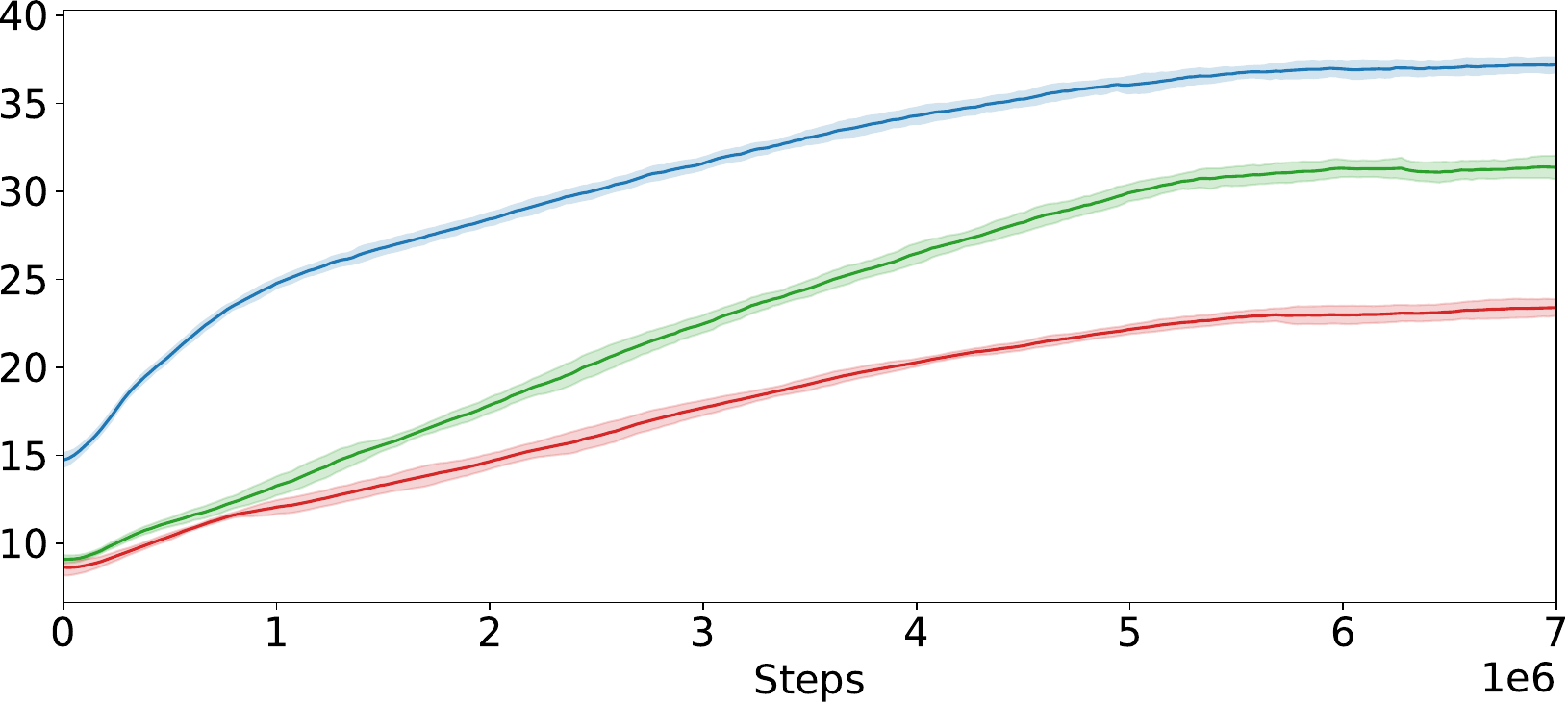} &
         \includegraphics[width=0.32\textwidth]{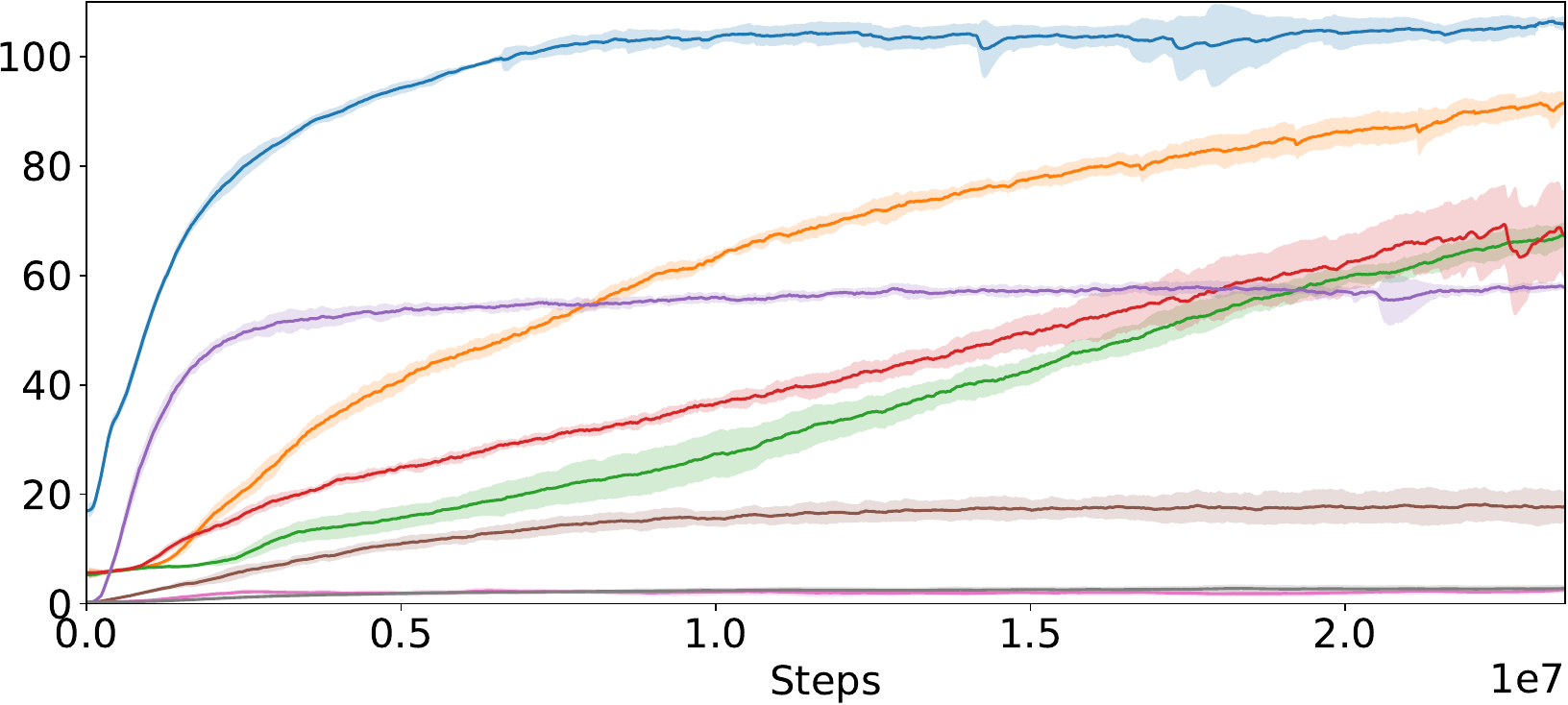}
    \end{tabular}
     \includegraphics[width=0.95\textwidth]{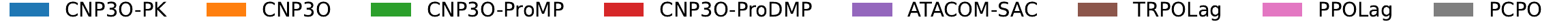}
     \vspace{-2mm}
    \caption{Learning curves (reward w.r.t. number of simulation steps) for the: (a) heavy object task without prior knowledge, (b) with prior knowledge, and (c) air hockey hitting task.}
    \label{fig:learning_curves}
\end{figure}
\vspace{-0.4cm}

\paragraph{Heavy object task} The first task consists of moving a heavy object with a manipulator from one block to another while avoiding collision, fulfilling torque limit requirements, and maintaining vertical orientation. In this task, kinodynamic constraints are particularly important due to the heavy mass mounted on the robot's end-effector, pushing the required torque commands to the limit.

The results of this experiment are reported in Figure~\ref{fig:learning_curves}ab.
The results show that our method outperforms all other approaches in both settings, independently of the use of prior knowledge. We could not obtain good learning performance with \gls{ppolag}, \gls{trpolag} and \gls{pcpo}. We believe that the combination of kinodynamic constraints and orientation constraints is too strict \revise{for both Lagrangian-style approach and projection-based one}.
Instead, the approaches using \glspl{mp} obtain satisfactory learning results. In particular, the B-spline representation (\gls{ours}) can obtain faster and more stable learning results than the alternative \glspl{mp}. We argue that the B-spline representation allows us to decouple the learning of the geometric path from the controlled execution's speed and, consequently, to deal more easily with different constraints with conflicting requirements.
Using \glspl{mp} is straightforward to introduce prior information into the task, e.g., force the target position for each planning setting. Exploiting prior information allows us to learn the task faster than relying on end-to-end methods. The prior information setting widens, even more, the performance gap between the B-Spline representation and other competing \gls{mp} parameterizations, presumably due to the ability to enforce zero joint velocity and acceleration at the end of the motion.

\begin{figure}[b!]
    \vspace{-1em}
    \centering
    \includegraphics[width=\textwidth]{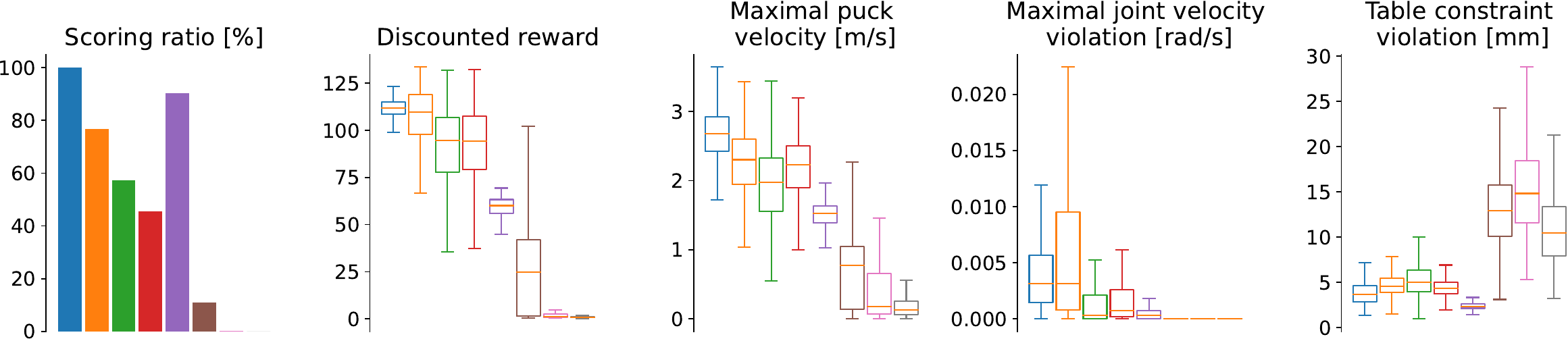}
    \includegraphics[width=0.95\textwidth]{imgs/legend.pdf}
    \vspace{-2mm}
    \caption{Statistical analysis of the considered approaches on the simulated Air Hockey hitting task.}
    \label{fig:air_hockey_sim}
\end{figure}

\paragraph{Robot Air Hockey Hitting} The second task is to teach a robot manipulator to score goals in the robot air hockey setting. 
Here, the constraints are mainly collision constraints preventing the robot from colliding with the table and the constraint of maintaining the end effector on the table surface. Notice that the policy must control all the joints of the robot, therefore maintaining the surface is extremely challenging without prior knowledge.

We can see the performance of the algorithms in Figure~\ref{fig:learning_curves}c. Here, it is clear that our methodology outperforms all the baselines. In particular, unsurprisingly, the \gls{ppolag}, \gls{trpolag} and \gls{pcpo} algorithms cannot achieve good performance. \revise{Also, in this case, we argue that this poor performance is mainly due to complex constraints that one side cannot be handled effectively without prior knowledge, and on the other side limit the exploration of the projection-based method.}
In turn, \gls{atacom} quickly converges to a suboptimal solution. This approach is much more suited for learning this task as it exploits prior knowledge. However, \gls{atacom} uses \gls{sac} as the underlying learning algorithm: stepwise exploration and automatic entropy tuning allow for very fast and effective learning; however, it is often prone to premature convergence.
In comparison, our learning method combined with \glspl{mp} converges to much more performant solutions. We argue this is due to the trajectory-based exploration in a lower-dimensional representation space, which allows more meaningful exploration and easy correction of the behaviors.
Regarding the trajectory representation, we also observe in this task that the more flexible B-splines approach allows our method to reach optimal results faster. Furthermore, by adding prior knowledge, our learning speed is comparable with \gls{atacom}, allowing us to reach higher values of the task objective, primarily due to faster-hitting behaviors.
We argue that if prior knowledge is fundamental for task safety, exploiting it to make learning more effective is straightforward, particularly if the framework allows easy encoding of the task information.

In Figure~\ref{fig:air_hockey_sim}, we present the detailed metrics of the simulated hitting behaviors at the end of the training. We evaluate the trajectories in terms of success rate, expected discounted return of the mean of the search distribution, maximum puck velocity, and constraint violations in terms of joint and end-effector violations. We observe that \gls{ours} achieves higher performance and less variant results in all the metrics. In particular, \gls{ours} with Prior information achieves a scoring ratio close to 100\%, and without prior information, only \gls{atacom} outperforms \gls{ours}. However, it is worth noting that the \gls{atacom} hitting is significantly slower. \revise{Moreover, none of the remaining \gls{saferl} baselines can get close to the \gls{ours} and \gls{atacom}, even they are allowed to make bigger constraints violations (see table constraint violations for \gls{ppolag}, \gls{trpolag} and \gls{pcpo} and Appendix~\ref{app:ah_task} for more detailed explanation)
Regarding constraints violations, all the \gls{mp} approaches and \gls{atacom} can maintain the end effector on the table surface quite accurately. }
The highest joint velocity violations are achieved by \gls{ours}-based methods, however, the scale of the violations is relatively small and it is primarily caused by the very fast motions at the edge of the robot capabilities.

\begin{figure}[t!]
    \centering
    \includegraphics[width=\textwidth]{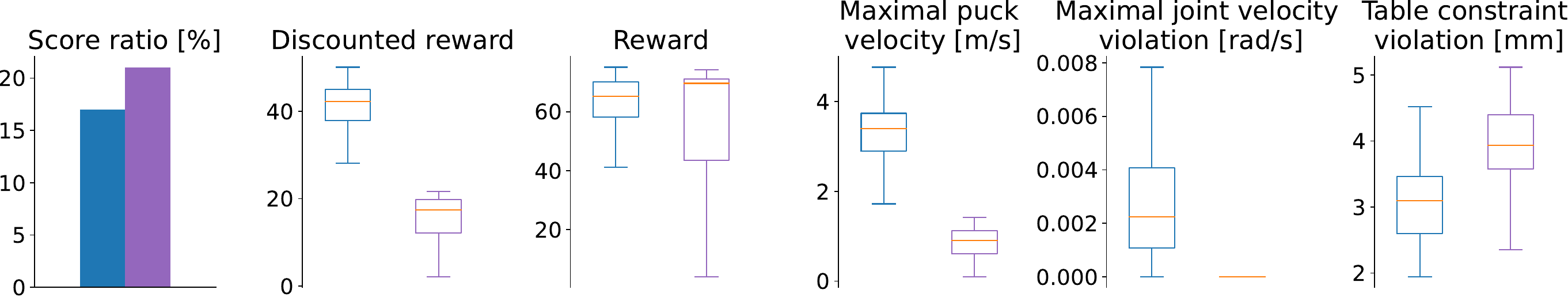}
    \includegraphics[width=0.3\textwidth]{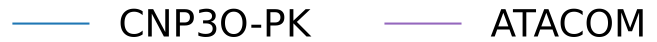}
    \vspace{-2mm}
    \caption{Statistical comparison of \gls{ours} and \gls{atacom} on the Air Hockey hitting with real robot.}
    \label{fig:air_hockey_real}
    \vspace{-0.2cm}
\end{figure}

Finally, we test the zero-shot capabilities of our approach in the real-world robot air hockey setting. We evaluate both the \gls{ours} and the \gls{atacom} policies for 100 hitting attempts. 
Looking at the metrics presented in Figure~\ref{fig:air_hockey_real}, we observe significant performance loss compared with the simulated task, particularly regarding success rate. This drop happens since our learning in the simulated environment does not employ any domain randomization technique. Thus, the big sim-to-real gap is due to the unmodelled disturbances of the air hockey table, imperfect controllers, and delays in both perception and command.
Nevertheless, the deployed policy can still hit strongly the puck and score some goals. 
Compared with \gls{atacom} zero shot deployment, we score fewer goals, however \gls{ours} achieves much higher total reward, mainly due to much faster hitting behaviors, as clearly shown in the boxplots. 
It is important to notice that our approach in the real world can obtain lower z-axis violations. We argue that this unexpected result can be attributed to the sampling of entire trajectories, instead of single actions at each timestep. Indeed,  this makes our approach less sensitive to the observation noise, which is effectively filtered by the low-level control loop.
Another issue for our method not so present in the behaviors learned by \gls{atacom}, is that the mallet occasionally flips during trajectory execution.
The metrics suggest that fast movements in combination with small z-axis violations cause this problem.

\paragraph{Limitations}
Our methodology has some limitations compared to standard \gls{saferl} algorithms. First of all, we require the knowledge of the constraints. However, the algorithm can be easily adapted to work with samples from the environment. 
The second limitation is that our approach is relatively more data-hungry than classical \gls{rl} approaches. This can be limited by
exploiting recent advances in the episodic \gls{rl}~\cite{li2024open}
and initializing the solution with imitation learning, speeding up the initial learning phases.
Furthermore, our algorithm for \gls{mp} learning is quite simple and may benefit from modern learning techniques~\citep{li2024open} for \gls{mp} or better trust-region approaches~\cite{otto2021differentiable}.
Finally, while it is possible to adapt the method to learn directly in the real world, this comes at the cost of being able to evaluate the full trajectories beforehand and preventing the execution of trajectories that violate safety constraints too much. 
Another alternative is to improve the zero-shot transfer of the policy on the robot. This could be done by including domain randomization in the training process or by adding replanning as done in~\cite{kicki2023fast}. However, both these approaches require even more samples for the training.


\section{Conclusion}
\label{sec:conclusion}
In this work, we bridge the gap between the \gls{saferl}, \gls{mp}, and learning-to-plan. The proposed solution combines the strengths of these approaches and outperforms state-of-the-art \gls{saferl} algorithms on two challenging robotic manipulation tasks with a variety of constraints. 
Our method utilizes the knowledge of the constraints and robot dynamics to improve learning efficiency but, unlike learning-to-plan methods, does not require full knowledge of the environment. Moreover, it leverages \glspl{mp} to reduce the dimensionality of the action space, facilitate reasonable exploration, and incorporate prior knowledge in the form of boundary conditions. Finally, the proposed solution learns how to generate trajectories satisfying a rich collection of constraints thanks to the introduced learning algorithm inspired by the techniques exploited in \gls{saferl}.

In our experiments, we show that \gls{ours} allows the robot to perform significantly more dynamic motions while ensuring only minimal violations of the constraints, comparable to much more conservative state-of-the-art approaches. Our real robot evaluation in robot air hockey shows the ability of the introduced method to maintain a safety level comparable to the one observed in simulation and transfer to complex real-world robotics tasks in a zero-shot manner.
Last but not least, we show that our proposed \gls{mp} parametrization is more suitable than existing \glspl{mp} in case of constrained optimization and also allows the designer to encode more prior knowledge in the learning process, such as desired configuration, velocity, acceleration, and higher order derivatives. This is particularly important for smoothly composing sequences of trajectories together to solve even more complex tasks, which we see as a potential direction for future research.



\clearpage
\acknowledgments{This paper is partially supported by the
German Federal Ministry of Education and Research (BMBF) within the
collaborative KIARA project (grant no. 13N16274)}


\bibliography{bibliography}  

\begin{thebibliography}{66}
\providecommand{\natexlab}[1]{#1}
\providecommand{\url}[1]{\texttt{#1}}
\expandafter\ifx\csname urlstyle\endcsname\relax
  \providecommand{\doi}[1]{doi: #1}\else
  \providecommand{\doi}{doi: \begingroup \urlstyle{rm}\Url}\fi

\bibitem[M{\"u}lling et~al.(2011)M{\"u}lling, Kober, and Peters]{mulling2011biomimetic}
K.~M{\"u}lling, J.~Kober, and J.~Peters.
\newblock A biomimetic approach to robot table tennis.
\newblock \emph{Adaptive Behavior}, 19\penalty0 (5):\penalty0 359--376, 2011.

\bibitem[B{\"u}chler et~al.(2022)B{\"u}chler, Guist, Calandra, Berenz, Sch{\"o}lkopf, and Peters]{buchler2022learning}
D.~B{\"u}chler, S.~Guist, R.~Calandra, V.~Berenz, B.~Sch{\"o}lkopf, and J.~Peters.
\newblock Learning to play table tennis from scratch using muscular robots.
\newblock \emph{IEEE Transactions on Robotics}, 2022.

\bibitem[Ploeger et~al.(2021)Ploeger, Lutter, and Peters]{ploeger2021high}
K.~Ploeger, M.~Lutter, and J.~Peters.
\newblock High acceleration reinforcement learning for real-world juggling with binary rewards.
\newblock In \emph{Conference on Robot Learning}, pages 642--653. PMLR, 2021.

\bibitem[Ploeger and Peters(2022)]{ploeger2022controlling}
K.~Ploeger and J.~Peters.
\newblock Controlling the cascade: Kinematic planning for n-ball toss juggling.
\newblock In \emph{2022 IEEE/RSJ International Conference on Intelligent Robots and Systems (IROS)}, pages 1139--1144. IEEE, 2022.

\bibitem[von Drigalski et~al.(2021)von Drigalski, Joshi, Murooka, Tanaka, Hamaya, and Ijiri]{von2021analytical}
F.~von Drigalski, D.~Joshi, T.~Murooka, K.~Tanaka, M.~Hamaya, and Y.~Ijiri.
\newblock An analytical diabolo model for robotic learning and control.
\newblock In \emph{2021 IEEE International Conference on Robotics and Automation (ICRA)}, pages 4055--4061. IEEE, 2021.

\bibitem[Zaidi et~al.(2023)Zaidi, Martin, Belles, Zakharov, Krishna, Lee, Wagstaff, Naik, Sklar, Choi, et~al.]{zaidi2023athletic}
Z.~Zaidi, D.~Martin, N.~Belles, V.~Zakharov, A.~Krishna, K.~M. Lee, P.~Wagstaff, S.~Naik, M.~Sklar, S.~Choi, et~al.
\newblock Athletic mobile manipulator system for robotic wheelchair tennis.
\newblock \emph{IEEE Robotics and Automation Letters}, 8\penalty0 (4):\penalty0 2245--2252, 2023.

\bibitem[Haarnoja et~al.(2024)Haarnoja, Moran, Lever, Huang, Tirumala, Humplik, Wulfmeier, Tunyasuvunakool, Siegel, Hafner, et~al.]{haarnoja2024learning}
T.~Haarnoja, B.~Moran, G.~Lever, S.~H. Huang, D.~Tirumala, J.~Humplik, M.~Wulfmeier, S.~Tunyasuvunakool, N.~Y. Siegel, R.~Hafner, et~al.
\newblock Learning agile soccer skills for a bipedal robot with deep reinforcement learning.
\newblock \emph{Science Robotics}, 9\penalty0 (89):\penalty0 eadi8022, 2024.

\bibitem[Altman(1998)]{altman1998constrained}
E.~Altman.
\newblock {Constrained Markov Decision Processes with Total Cost Criteria: Lagrangian Approach and Dual Linear Program}.
\newblock \emph{Mathematical methods of operations research}, 48\penalty0 (3):\penalty0 387--417, 1998.

\bibitem[Kicki et~al.(2023)Kicki, Liu, Tateo, Bou-Ammar, Walas, Skrzypczy{\'n}ski, and Peters]{kicki2023fast}
P.~Kicki, P.~Liu, D.~Tateo, H.~Bou-Ammar, K.~Walas, P.~Skrzypczy{\'n}ski, and J.~Peters.
\newblock Fast kinodynamic planning on the constraint manifold with deep neural networks.
\newblock \emph{IEEE Transactions on Robotics}, 2023.

\bibitem[Altman(1999)]{altman1999constrained}
E.~Altman.
\newblock \emph{Constrained Markov decision processes: stochastic modeling}.
\newblock Routledge, 1999.

\bibitem[Tessler et~al.(2019)Tessler, Mankowitz, and Mannor]{tessler2019reward}
C.~Tessler, D.~J. Mankowitz, and S.~Mannor.
\newblock {Reward Constrained Policy Optimization}.
\newblock In \emph{International Conference on Learning Representations (ICLR)}, 2019.

\bibitem[Stooke et~al.(2020)Stooke, Achiam, and Abbeel]{stooke2020responsive}
A.~Stooke, J.~Achiam, and P.~Abbeel.
\newblock {Responsive Safety in Reinforcement Learning by PID Lagrangian Methods}.
\newblock In \emph{International Conference on Machine Learning (ICML)}, 2020.

\bibitem[Gangapurwala et~al.(2020)Gangapurwala, Mitchell, and Havoutis]{ioannis2022guided}
S.~Gangapurwala, A.~Mitchell, and I.~Havoutis.
\newblock Guided constrained policy optimization for dynamic quadrupedal robot locomotion.
\newblock \emph{IEEE Robotics and Automation Letters}, 5\penalty0 (2):\penalty0 3642--3649, 2020.
\newblock \doi{10.1109/LRA.2020.2979656}.

\bibitem[Ding et~al.(2021)Ding, Wei, Yang, Wang, and Jovanovic]{ding2021provably}
D.~Ding, X.~Wei, Z.~Yang, Z.~Wang, and M.~R. Jovanovic.
\newblock {Provably Efficient Safe Exploration via Primal-Dual Policy Optimization}.
\newblock In \emph{International Conference on Artificial Intelligence and Statistics (AISTATS)}, volume 130, 2021.

\bibitem[Borkar and Jain(2014)]{borkar2014risk}
V.~Borkar and R.~Jain.
\newblock Risk-constrained markov decision processes.
\newblock \emph{IEEE Transactions on Automatic Control}, 59\penalty0 (9):\penalty0 2574--2579, 2014.

\bibitem[Ying et~al.(2022)Ying, Zhou, Su, Yan, Chen, and Zhu]{ying2022towards}
C.~Ying, X.~Zhou, H.~Su, D.~Yan, N.~Chen, and J.~Zhu.
\newblock Towards safe reinforcement learning via constraining conditional value-at-risk.
\newblock In \emph{International Joint Conference on Artificial Intelligence}, 2022.

\bibitem[Yang et~al.(2023)Yang, Sim{\~a}o, Tindemans, and Spaan]{yang2023safety}
Q.~Yang, T.~D. Sim{\~a}o, S.~H. Tindemans, and M.~T. Spaan.
\newblock Safety-constrained reinforcement learning with a distributional safety critic.
\newblock \emph{Machine Learning}, 112\penalty0 (3):\penalty0 859--887, 2023.

\bibitem[Sootla et~al.(2022)Sootla, Cowen-Rivers, Wang, and Ammar]{sootla2022enhancing}
A.~Sootla, A.~I. Cowen-Rivers, J.~Wang, and H.~B. Ammar.
\newblock Enhancing safe exploration using safety state augmentation.
\newblock In A.~H. Oh, A.~Agarwal, D.~Belgrave, and K.~Cho, editors, \emph{Advances in Neural Information Processing Systems}, 2022.

\bibitem[Achiam et~al.(2017)Achiam, Held, Tamar, and Abbeel]{achiam2017constrained}
J.~Achiam, D.~Held, A.~Tamar, and P.~Abbeel.
\newblock {Constrained Policy Optimization}.
\newblock In \emph{International Conference on Machine Learning (ICML)}, 2017.

\bibitem[Kim and Oh(2022)]{kim2022efficient}
D.~Kim and S.~Oh.
\newblock Efficient off-policy safe reinforcement learning using trust region conditional value at risk.
\newblock \emph{IEEE Robotics and Automation Letters}, 7\penalty0 (3):\penalty0 7644--7651, 2022.

\bibitem[Liu et~al.(2020)Liu, Ding, and Liu]{liu2020ipo}
Y.~Liu, J.~Ding, and X.~Liu.
\newblock Ipo: Interior-point policy optimization under constraints.
\newblock In \emph{AAAI Conference on Artificial Intelligence (AAAI)}, volume 34(04), pages 4940--4947, 2020.

\bibitem[Chow et~al.(2018)Chow, Nachum, Duenez-Guzman, and Ghavamzadeh]{chow2018lyapunov}
Y.~Chow, O.~Nachum, E.~Duenez-Guzman, and M.~Ghavamzadeh.
\newblock {A Lyapunov-based Approach to Safe Reinforcement Learning}.
\newblock In \emph{Conference on Neural Information Processing Systems (NIPS)}, 2018.

\bibitem[Chow et~al.(2019)Chow, Nachum, Faust, Duenez-Guzman, and Ghavamzadeh]{chow2019lyapunov}
Y.~Chow, O.~Nachum, A.~Faust, E.~Duenez-Guzman, and M.~Ghavamzadeh.
\newblock {Lyapunov-based Safe Policy Optimization for Continuous Control}.
\newblock In \emph{Reinforcement Learning for Real Life (RL4RealLife) Workshop in the 36 th International Conference on Machine Learning}, 2019.

\bibitem[Sikchi et~al.(2021)Sikchi, Zhou, and Held]{sikchi2021lyapunov}
H.~Sikchi, W.~Zhou, and D.~Held.
\newblock Lyapunov barrier policy optimization.
\newblock \emph{arXiv preprint arXiv:2103.09230}, 2021.

\bibitem[Ames et~al.(2019)Ames, Coogan, Egerstedt, Notomista, Sreenath, and Tabuada]{ames2019control}
A.~D. Ames, S.~Coogan, M.~Egerstedt, G.~Notomista, K.~Sreenath, and P.~Tabuada.
\newblock Control barrier functions: Theory and applications.
\newblock In \emph{2019 18th European control conference (ECC)}, pages 3420--3431. IEEE, 2019.

\bibitem[Xiao and Belta(2022)]{xiao2022high_order}
W.~Xiao and C.~Belta.
\newblock High-{Order} {Control} {Barrier} {Functions}.
\newblock \emph{IEEE Transactions on Automatic Control}, 67\penalty0 (7):\penalty0 3655--3662, July 2022.
\newblock ISSN 1558-2523.
\newblock Conference Name: IEEE Transactions on Automatic Control.

\bibitem[Taylor et~al.(2020)Taylor, Singletary, Yue, and Ames]{taylor2020learning}
A.~Taylor, A.~Singletary, Y.~Yue, and A.~Ames.
\newblock Learning for safety-critical control with control barrier functions.
\newblock In \emph{Learning for Dynamics and Control}, pages 708--717. PMLR, 2020.

\bibitem[Cheng et~al.(2019)Cheng, Orosz, Murray, and Burdick]{cheng2019end}
R.~Cheng, G.~Orosz, R.~M. Murray, and J.~W. Burdick.
\newblock {End-to-End Safe Reinforcement Learning through Barrier Functions for Safety-Critical Continuous Control Tasks}.
\newblock In \emph{AAAI Conference on Artificial Intelligence}, pages 3387--3395. AAAI Press, 2019.

\bibitem[Fisac et~al.(2018)Fisac, Akametalu, Zeilinger, Kaynama, Gillula, and Tomlin]{fisac2018general}
J.~F. Fisac, A.~K. Akametalu, M.~N. Zeilinger, S.~Kaynama, J.~Gillula, and C.~J. Tomlin.
\newblock A general safety framework for learning-based control in uncertain robotic systems.
\newblock \emph{IEEE Transactions on Automatic Control}, 64\penalty0 (7):\penalty0 2737--2752, 2018.

\bibitem[Shao et~al.(2021)Shao, Chen, Kousik, and Vasudevan]{shao2021reachability}
Y.~S. Shao, C.~Chen, S.~Kousik, and R.~Vasudevan.
\newblock Reachability-based trajectory safeguard (rts): A safe and fast reinforcement learning safety layer for continuous control.
\newblock \emph{IEEE Robotics and Automation Letters}, 6\penalty0 (2):\penalty0 3663--3670, 2021.

\bibitem[Selim et~al.(2022)Selim, Alanwar, Kousik, Gao, Pavone, and Johansson]{selim2022safe}
M.~Selim, A.~Alanwar, S.~Kousik, G.~Gao, M.~Pavone, and K.~H. Johansson.
\newblock Safe reinforcement learning using black-box reachability analysis.
\newblock \emph{IEEE Robotics and Automation Letters}, 7\penalty0 (4):\penalty0 10665--10672, 2022.

\bibitem[Zheng et~al.(2024)Zheng, Li, Yu, Yang, Li, Zhan, and Liu]{zheng2024safe}
Y.~Zheng, J.~Li, D.~Yu, Y.~Yang, S.~E. Li, X.~Zhan, and J.~Liu.
\newblock Safe offline reinforcement learning with feasibility-guided diffusion model.
\newblock In \emph{The Twelfth International Conference on Learning Representations}, 2024.

\bibitem[Hans et~al.(2008)Hans, Schneega{\ss}, Sch{\"{a}}fer, and Udluft]{hans2008safe}
A.~Hans, D.~Schneega{\ss}, A.~M. Sch{\"{a}}fer, and S.~Udluft.
\newblock {Safe Exploration for Reinforcement Learning}.
\newblock In \emph{European Symposium on Artificial Neural Networks (ESANN)}, 2008.

\bibitem[Garcia and Fernandez(2012)]{garcia2012safe}
J.~Garcia and F.~Fernandez.
\newblock {Safe Exploration of State and Action Spaces in Reinforcement Learning}.
\newblock \emph{Journal of Artificial Intelligence Research}, 45:\penalty0 515--564, 2012.
\newblock ISSN 10769757.

\bibitem[Berkenkamp et~al.(2017)Berkenkamp, Turchetta, Schoellig, and Krause]{berkenkamp2017safe}
F.~Berkenkamp, M.~Turchetta, A.~P. Schoellig, and A.~Krause.
\newblock {Safe Model-based Reinforcement Learning with Stability Guarantees}.
\newblock In \emph{Conference on Neural Information Processing Systems (NIPS)}, 2017.

\bibitem[Fisac et~al.(2019)Fisac, Lugovoy, Rubies-Royo, Ghosh, and Tomlin]{fisac2019bridging}
J.~F. Fisac, N.~F. Lugovoy, V.~Rubies-Royo, S.~Ghosh, and C.~J. Tomlin.
\newblock Bridging hamilton-jacobi safety analysis and reinforcement learning.
\newblock In \emph{2019 International Conference on Robotics and Automation (ICRA)}, pages 8550--8556. IEEE, 2019.

\bibitem[Pham et~al.(2018)Pham, De~Magistris, and Tachibana]{pham2018optlayer}
T.-H. Pham, G.~De~Magistris, and R.~Tachibana.
\newblock Optlayer-practical constrained optimization for deep reinforcement learning in the real world.
\newblock In \emph{2018 IEEE International Conference on Robotics and Automation (ICRA)}, pages 6236--6243. IEEE, 2018.

\bibitem[Dalal et~al.(2018)Dalal, Dvijotham, Vecerik, Hester, Paduraru, and Tassa]{dalal2018safe}
G.~Dalal, K.~Dvijotham, M.~Vecerik, T.~Hester, C.~Paduraru, and Y.~Tassa.
\newblock {Safe Exploration in Continuous Action Spaces}.
\newblock \emph{arXiv preprint arXiv:1801.08757}, 2018.

\bibitem[Liu et~al.(2022)Liu, Tateo, Ammar, and Peters]{liu2022robot}
P.~Liu, D.~Tateo, H.~B. Ammar, and J.~Peters.
\newblock Robot {Reinforcement} {Learning} on the {Constraint} {Manifold}.
\newblock In \emph{Conference on {Robot} {Learning}}, pages 1357--1366. PMLR, 2022.

\bibitem[Emam et~al.(2022)Emam, Notomista, Glotfelter, Kira, and Egerstedt]{emam2022safe}
Y.~Emam, G.~Notomista, P.~Glotfelter, Z.~Kira, and M.~Egerstedt.
\newblock Safe reinforcement learning using robust control barrier functions.
\newblock \emph{IEEE Robotics and Automation Letters}, 2022.

\bibitem[Liu et~al.(2023)Liu, Zhang, Tateo, Jauhri, Hu, Peters, and Chalvatzaki]{liu2023safe}
P.~Liu, K.~Zhang, D.~Tateo, S.~Jauhri, Z.~Hu, J.~Peters, and G.~Chalvatzaki.
\newblock Safe {Reinforcement} {Learning} of {Dynamic} {High}-{Dimensional} {Robotic} {Tasks}: {Navigation}, {Manipulation}, {Interaction}.
\newblock In \emph{Proceedings of the {IEEE} {International} {Conference} on {Robotics} and {Automation}}. IEEE, 2023.

\bibitem[LaValle(2006)]{lavalle2006planning}
S.~M. LaValle.
\newblock \emph{Planning Algorithms}.
\newblock Cambridge University Press, USA, 2006.
\newblock ISBN 0521862051.

\bibitem[Qureshi et~al.(2020)Qureshi, Miao, Simeonov, and Yip]{qureshi2019mpnet}
A.~H. Qureshi, Y.~Miao, A.~Simeonov, and M.~C. Yip.
\newblock Motion planning networks: Bridging the gap between learning-based and classical motion planners.
\newblock \emph{IEEE Transactions on Robotics}, pages 1--9, 2020.

\bibitem[Kicki and Skrzypczyński(2022)]{kicki2022speedingup}
P.~Kicki and P.~Skrzypczyński.
\newblock Speeding up deep neural network-based planning of local car maneuvers via efficient b-spline path construction.
\newblock In \emph{2022 International Conference on Robotics and Automation (ICRA)}, pages 4422--4428, 2022.

\bibitem[Johnson et~al.(2023)Johnson, Qureshi, and Yip]{johnson2023transformers}
J.~J. Johnson, A.~H. Qureshi, and M.~C. Yip.
\newblock Learning sampling dictionaries for efficient and generalizable robot motion planning with transformers.
\newblock \emph{IEEE Robotics and Automation Letters}, 8\penalty0 (12):\penalty0 7946--7953, 2023.

\bibitem[Carvalho et~al.(2023)Carvalho, Le, Baierl, Koert, and Peters]{joao2023diffusion}
J.~Carvalho, A.~T. Le, M.~Baierl, D.~Koert, and J.~Peters.
\newblock Motion planning diffusion: Learning and planning of robot motions with diffusion models.
\newblock In \emph{2023 IEEE/RSJ International Conference on Intelligent Robots and Systems (IROS)}, pages 1916--1923, 2023.

\bibitem[Kicki et~al.(2021)Kicki, Gawron, Ćwian, Ozay, and Skrzypczyński]{kicki2021eaai}
P.~Kicki, T.~Gawron, K.~Ćwian, M.~Ozay, and P.~Skrzypczyński.
\newblock Learning from experience for rapid generation of local car maneuvers.
\newblock \emph{Engineering Applications of Artificial Intelligence}, 105:\penalty0 104399, 2021.
\newblock ISSN 0952-1976.

\bibitem[Osa(2022)]{osa2022mp}
T.~Osa.
\newblock Motion planning by learning the solution manifold in trajectory optimization.
\newblock \emph{The International Journal of Robotics Research}, 41\penalty0 (3):\penalty0 281--311, 2022.

\bibitem[Paraschos et~al.(2013)Paraschos, Daniel, Peters, and Neumann]{promp}
A.~Paraschos, C.~Daniel, J.~R. Peters, and G.~Neumann.
\newblock Probabilistic movement primitives.
\newblock In C.~Burges, L.~Bottou, M.~Welling, Z.~Ghahramani, and K.~Weinberger, editors, \emph{Advances in Neural Information Processing Systems}, volume~26. Curran Associates, Inc., 2013.

\bibitem[Li et~al.(2023)Li, Jin, Volpp, Otto, Lioutikov, and Neumann]{li2023prodmp}
G.~Li, Z.~Jin, M.~Volpp, F.~Otto, R.~Lioutikov, and G.~Neumann.
\newblock Prodmp: A unified perspective on dynamic and probabilistic movement primitives.
\newblock \emph{IEEE Robotics and Automation Letters}, 8\penalty0 (4):\penalty0 2325--2332, 2023.

\bibitem[Saveriano et~al.(2023)Saveriano, Abu-Dakka, Kramberger, and Peternel]{dmp}
M.~Saveriano, F.~J. Abu-Dakka, A.~Kramberger, and L.~Peternel.
\newblock Dynamic movement primitives in robotics: A tutorial survey.
\newblock \emph{The International Journal of Robotics Research}, 42\penalty0 (13):\penalty0 1133--1184, 2023.

\bibitem[Lee et~al.(2023)Lee, Lee, Kim, Son, and Park]{lee2023equivariant}
B.~Lee, Y.~Lee, S.~Kim, M.~Son, and F.~C. Park.
\newblock Equivariant motion manifold primitives.
\newblock In \emph{7th Annual Conference on Robot Learning}, 2023.

\bibitem[Liu et~al.(2024)Liu, Bou-Ammar, Peters, and Tateo]{liu2024safe}
P.~Liu, H.~Bou-Ammar, J.~Peters, and D.~Tateo.
\newblock Safe reinforcement learning on the constraint manifold: Theory and applications.
\newblock \emph{arXiv preprint arXiv:2404.09080}, 2024.

\bibitem[Achiam and Amodei(2019)]{achiam2019benchmarking}
J.~Achiam and D.~Amodei.
\newblock Benchmarking safe exploration in deep reinforcement learning, 2019.

\bibitem[Yang et~al.(2020)Yang, Rosca, Narasimhan, and Ramadge]{tsung2020pcpo}
T.~Yang, J.~Rosca, K.~Narasimhan, and P.~J. Ramadge.
\newblock Projection-based constrained policy optimization.
\newblock In \emph{8th International Conference on Learning Representations, {ICLR} 2020, Addis Ababa, Ethiopia, April 26-30, 2020}. OpenReview.net, 2020.
\newblock URL \url{https://openreview.net/forum?id=rke3TJrtPS}.

\bibitem[Li et~al.(2024)Li, Zhou, Roth, Thilges, Otto, Lioutikov, and Neumann]{li2024open}
G.~Li, H.~Zhou, D.~Roth, S.~Thilges, F.~Otto, R.~Lioutikov, and G.~Neumann.
\newblock Open the black box: Step-based policy updates for temporally-correlated episodic reinforcement learning.
\newblock In \emph{The Twelfth International Conference on Learning Representations}, 2024.

\bibitem[Otto et~al.(2021)Otto, Becker, Ngo, Ziesche, and Neumann]{otto2021differentiable}
F.~Otto, P.~Becker, V.~A. Ngo, H.~C.~M. Ziesche, and G.~Neumann.
\newblock Differentiable trust region layers for deep reinforcement learning.
\newblock In \emph{International Conference on Learning Representations}, 2021.

\bibitem[Todorov et~al.(2012)Todorov, Erez, and Tassa]{mujoco}
E.~Todorov, T.~Erez, and Y.~Tassa.
\newblock Mujoco: A physics engine for model-based control.
\newblock In \emph{2012 IEEE/RSJ International Conference on Intelligent Robots and Systems}, pages 5026--5033, 2012.

\bibitem[Liu et~al.(2023)Liu, Guenster, and Tateo]{ahc}
P.~Liu, J.~Guenster, and D.~Tateo.
\newblock Air hockey challenge.
\newblock https://github.com/AirHockeyChallenge/air\_hockey\_challenge/tree/tournament, 2023.

\bibitem[Liu et~al.(2021)Liu, Tateo, Bou-Ammar, and Peters]{liu2021efficient}
P.~Liu, D.~Tateo, H.~Bou-Ammar, and J.~Peters.
\newblock Efficient and reactive planning for high speed robot air hockey.
\newblock In \emph{2021 IEEE/RSJ International Conference on Intelligent Robots and Systems (IROS)}, pages 586--593. IEEE, 2021.

\bibitem[Muelling et~al.(2010)Muelling, Kober, and Peters]{muelling2010learning}
K.~Muelling, J.~Kober, and J.~Peters.
\newblock Learning table tennis with a mixture of motor primitives.
\newblock In \emph{2010 10th IEEE-RAS international conference on humanoid robots}, pages 411--416. IEEE, 2010.

\bibitem[Maeda et~al.(2014)Maeda, Ewerton, Lioutikov, Amor, Peters, and Neumann]{maeda2014learning}
G.~Maeda, M.~Ewerton, R.~Lioutikov, H.~B. Amor, J.~Peters, and G.~Neumann.
\newblock Learning interaction for collaborative tasks with probabilistic movement primitives.
\newblock In \emph{2014 IEEE-RAS International Conference on Humanoid Robots}, pages 527--534. IEEE, 2014.

\bibitem[Parisi et~al.(2015)Parisi, Abdulsamad, Paraschos, Daniel, and Peters]{parisi2015reinforcement}
S.~Parisi, H.~Abdulsamad, A.~Paraschos, C.~Daniel, and J.~Peters.
\newblock Reinforcement learning vs human programming in tetherball robot games.
\newblock In \emph{2015 IEEE/RSJ International Conference on Intelligent Robots and Systems (IROS)}, pages 6428--6434. IEEE, 2015.

\bibitem[Celik et~al.(2022)Celik, Zhou, Li, Becker, and Neumann]{celik2022specializing}
O.~Celik, D.~Zhou, G.~Li, P.~Becker, and G.~Neumann.
\newblock Specializing versatile skill libraries using local mixture of experts.
\newblock In \emph{Conference on Robot Learning}, pages 1423--1433. PMLR, 2022.

\bibitem[Otto et~al.(2023)Otto, Celik, Zhou, Ziesche, Ngo, and Neumann]{otto2023deep}
F.~Otto, O.~Celik, H.~Zhou, H.~Ziesche, V.~A. Ngo, and G.~Neumann.
\newblock Deep black-box reinforcement learning with movement primitives.
\newblock In \emph{Conference on Robot Learning}, pages 1244--1265. PMLR, 2023.

\bibitem[Drolet et~al.(2023)Drolet, Campbell, and Amor]{drolet2023learning}
M.~Drolet, J.~Campbell, and H.~B. Amor.
\newblock Learning and blending robot hugging behaviors in time and space.
\newblock In \emph{2023 IEEE International Conference on Robotics and Automation (ICRA)}, pages 12071--12077. IEEE, 2023.

\end{thebibliography}

\clearpage

\appendix

\section*{Appendix}

\section{Environments}
In this section, we would like to introduce the details of the environments used to evaluate the method introduced in the paper and the baselines. Both of the considered simulation environments were implemented using the MuJoCo physics simulator~\cite{mujoco}.

\subsection{Heavy Object}
\begin{figure}[b!]
    \centering
    \begin{tabular}{cc}
         \scriptsize \textbf{Initial state} & \scriptsize \textbf{End state}\\
         \includegraphics[width=0.49\textwidth]{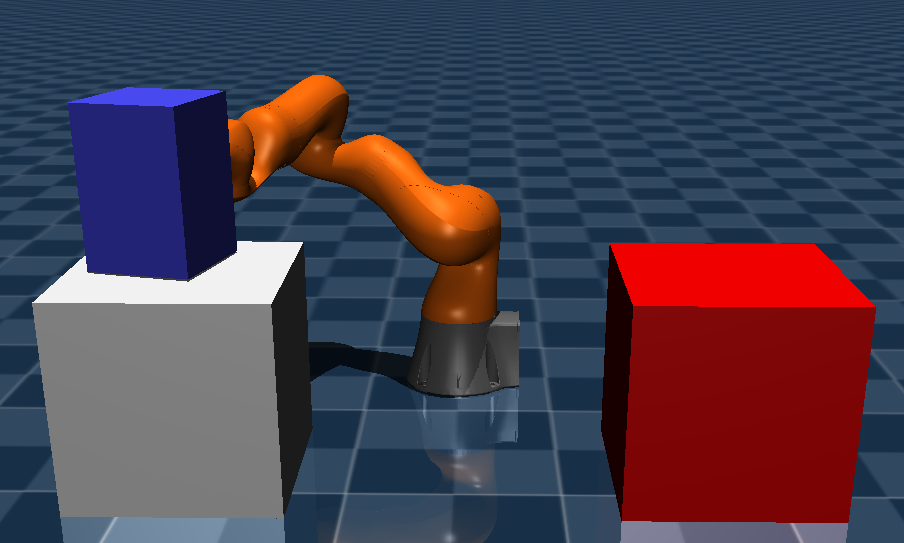} & \includegraphics[width=0.49\textwidth]{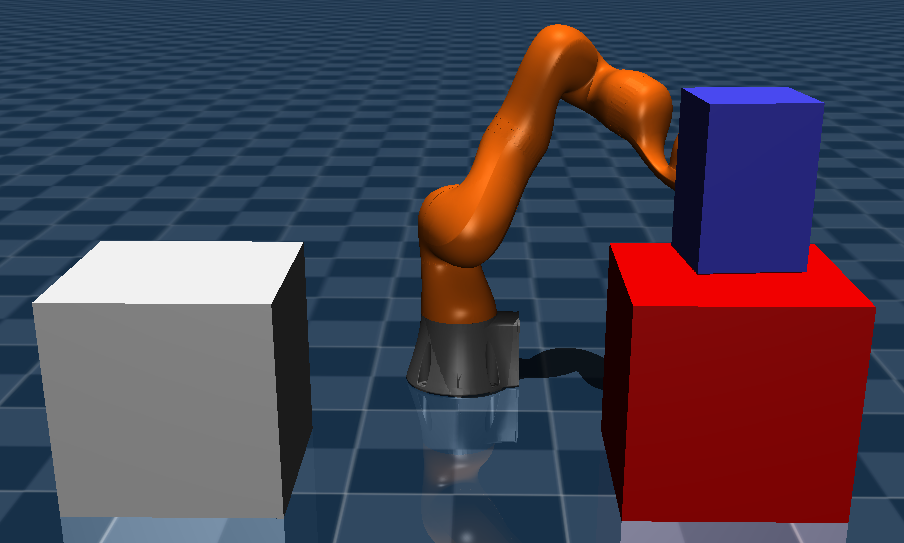}
    \end{tabular}
    \caption{Visualization of the task of moving a heavy object.}
    \label{fig:heavy_object_env}
\end{figure}

The Heavy Object environment is heavily inspired by the motion planning task of moving a heavy vertically oriented object introduced in~\cite{kicki2023fast}. In this task, the objective is to control a Kuka Iiwa 14 manipulator holding a heavy box weighing 12kg, such that the box, which is initially placed on one pedestal, is moved to some position on the second pedestal. The main difficulty of this task stems from the fact that the box is pushing the manipulator to its payload limit, which may result in exceeding the maximal torque that can be applied to the robot's joints. We make this task even more challenging by (i) adding constraints on the vertical orientation of the handled object, (ii) preferring faster motions due to the discount factor $\gamma = 0.99$, and (iii) minimizing the sum of the torques applied.
Finally, we require both the robot and the object to not collide with the pedestals.
A visualization of this task is presented in Figure~\ref{fig:heavy_object_env}.

A single episode is defined by the initial joint configuration of the robot, which is computed by the inverse kinematics based on the randomly drawn initial position of the handled object placed on top of the first pedestal, and the desired end position of the object placed on top of the second pedestal. The episode horizon is set to 100 steps, each consisting of 20 intermediate 1ms-long steps, which gives 2 s for the whole episode. 

In this task, the reward function is meant to encourage minimizing the distance to the goal pose, stopping the robot after reaching the goal, and not using too much energy. To achieve this we proposed the following reward function:
\begin{align}
    \scalemath{0.9}{R = \frac{1}{10 d + 1} + \text{I}(d < 0.01) \frac{0.01}{\|\dot{q}\| + 0.01} - 10^{-6} \|\bm{m}\|^2},
\end{align}
where $d$ is the Euclidean distance between the current and desired position of the heavy object, $\text{I}$ is an indicator function that is $1$ if the argument is true and $0$ otherwise, while $\bm{m}$ is the vector of joint torques.

Besides the reward function, we also defined several constraints, which are presented in Table~\ref{tab:kino_constraints}. The vertical orientation constraint is defined by the element $(2,2)$ of the handled object rotation matrix $R^{o}$. In turn, the collision loss is the sum of the collision between all considered collision bodies. We consider collisions between the robot body and pedestals, as well as, the handled heavy object and pedestals. For simplicity, we approximate the robot body by a sequence of 14cm radius balls located along the kinematic chain, no further than 10cm apart from each other. The value of the $\text{collision}(i, j)$ is the analytically computed depth of the penetration.
Moreover, there is an implicit constraint imposed on the maximal torques implemented by saturation of the control signals that are possible to be applied to the environment.

\paragraph{Task definition.} \revise{As explained in the main paper, in our setting we assume to have access to a task definition vector $T$. To fully describe this manipulation task, the task description vector contains the initial position and velocity of the robot and the desired pose of the handled object. Notice that this information is the minimal required set to perform the desired motion in a multitask setting.}

\begin{table}[h!]
    \centering
    \caption{Definition of the constraints in the Heavy Object task.}
    \begin{tabular}{c|c|c}
    No. & Name & Definition \\
    \hline
    1-7 & Joint positions & $|q| \leq [2.97, 2.09 , 2.97, 2.09 , 2.97, 2.09 , 3.05]$\\
    8-14 & Joint velocities & $|\dot{q}| \leq [1.48, 1.48, 1.75, 1.31, 2.27, 2.36, 2.36]$\\
    15 & Vertical orientation & $1 - R_{2,2}^{ho}$\\
    16 & Collisions & $\sum_{i, j} \text{collision}(i, j)$\\
    \end{tabular}
    \label{tab:kino_constraints}
\end{table}

\subsection{Air Hockey Hitting}
\label{app:ah_task}
\begin{figure}[t]
    \centering
    \includegraphics[height=3cm]{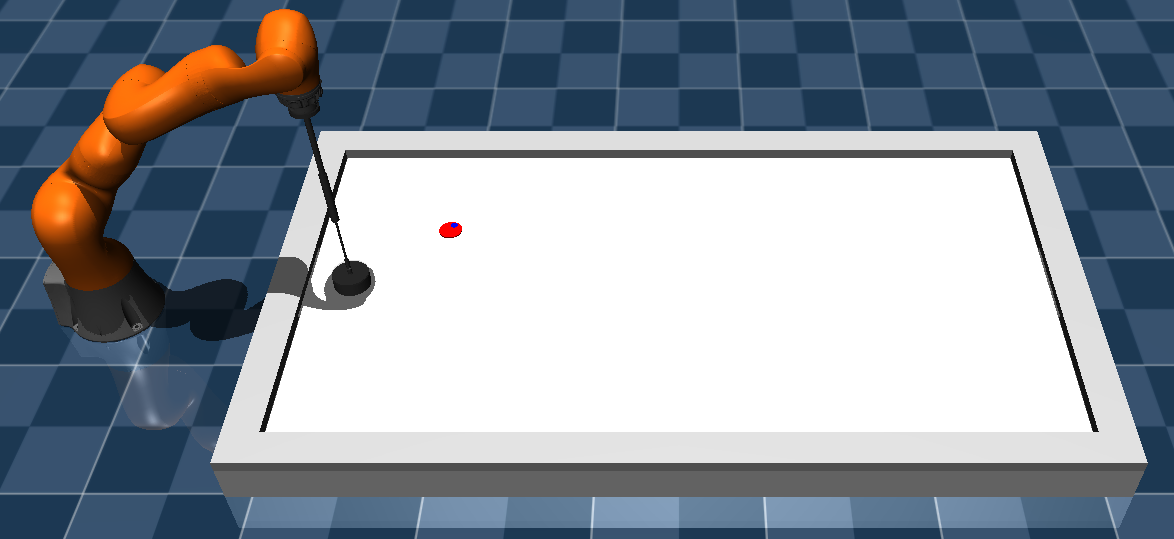}
    \includegraphics[height=3cm]{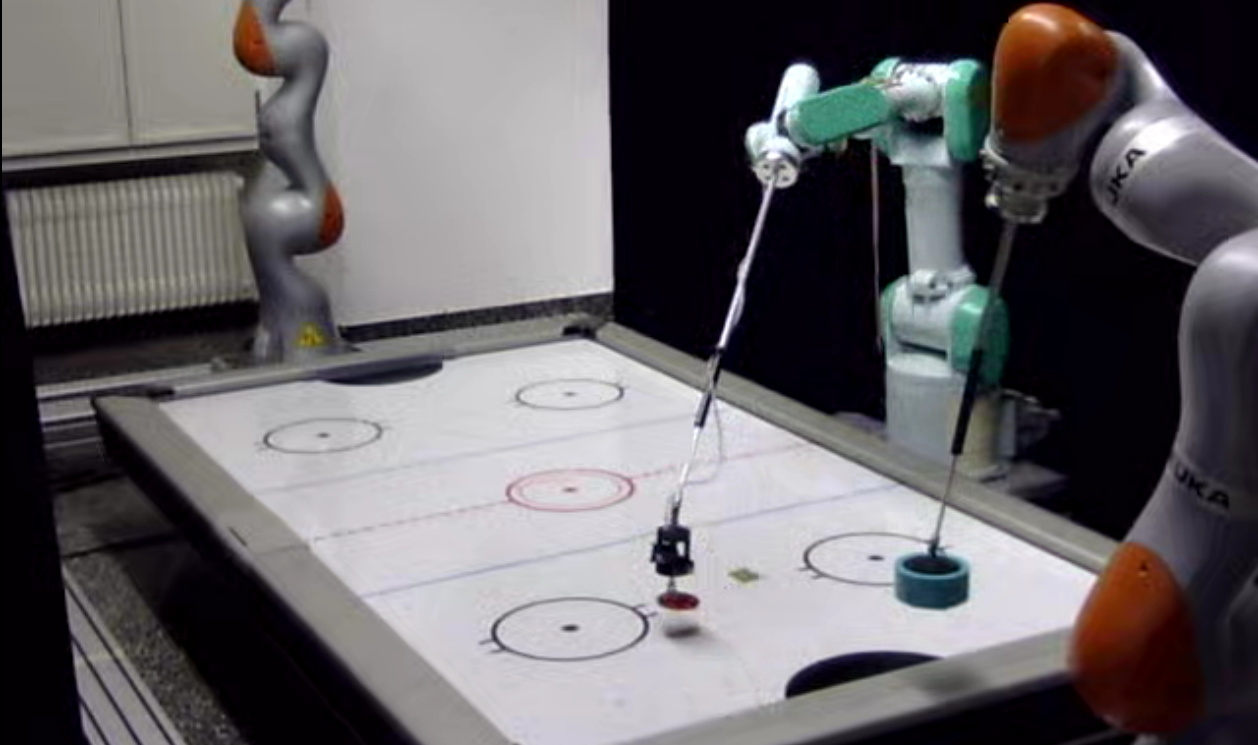}
    \caption{Visualization of the simulated and real-world Air Hockey Hitting task.}
    \label{fig:air_hockey_env}
\end{figure}

In this environment, the main objective is to hit the air hockey puck located on the air hockey table in such a way that it reaches the opponent's goal. The main difficulty of this task comes from the use of a general-purpose 7DoF manipulator (Kuka Iiwa 14) with a long end-effector ending with a mallet. This is particularly challenging due to the constraints that are put on the end-effector, i.e. remaining on the table plane throughout the whole movement, and the requirement to hit as fast as possible and, at the same time, very accurate.
The considered setup is presented in Figure~\ref{fig:air_hockey_env}.

The considered Air Hockey Hitting environment is a slightly adjusted version of the environment \textit{AirHockeyHit} introduced in the Air Hockey Challenge~\cite{ahc}.
We kept the original environment in an unchanged form, except for the slightly bigger set of puck initial positions ($[-0.65, -0.25] \times [-0.4, 0.4] \rightarrow [-0.7, -0.2] \times [-0.35, 0.35]$), smaller initial puck velocities range ($[0, 0.5] \rightarrow [0, 0.3]$), initial robot configuration fixed to a single one ($q_0 = [0, 0.697, 0, -0.505, 0, 1.929, 0]$) and shorter episode horizon ($500 \rightarrow 150$ steps).
Also, the reward function is the same as in the \textit{AirHockeyHit} environment. For non-absorbing states it gives a reward of $1.5 \clip(\dot{x}_p, 0, 3)$, where $\dot{x}_p$ is the puck velocity in $x$ axis, if the puck is in the opponents half of the table. It also encourages the robot's end-effector to get closer to the puck, by rewarding decrease of the distance between them multiplied by a factor of 10, if they were never that close before. Moreover, depending on the type of the absorbing state a different reward, scaled by the $\frac{1 - \gamma^h}{1 - \gamma}$, where $h$ is the environment horizon, is given.
In case of scoring the goal a reward $r=1.5 - 5 \clip(|y_p|, 0, 0.1)$ is awarded, for reaching the opponent band $r = 2 (1 - 2 \clip(|y_p| - 1, 0, 0.35))$, and in case of hitting the left or right band (from the player perspective) $r = 0.3 - 0.3 \clip(l - x_p, 0, 1)$, where $(x_p, y_p)$ is the puck position and $l$ is the length of the table.

Also in the case of this task, we require the satisfaction of both joint position and joint velocities limits and impose the torque constraints by restricting the actuation range. Besides them, we introduce constraints that stem from the table geometry, i.e. avoiding hitting left, right, and robot's own band (the opponent's band is not reachable) and remaining on the table surface. Exact definition of all of these constraints is provided in Table~\ref{tab:ah_constraints}, where $(x_{ee}, y_{ee}, z_{ee})$ is the end-effector position, $x_{ab}, y_{lb}, y_{rb}$ are the $x$ and $y$ coordinates of the robot's, left and right bands respectively, and $z_t$ is the $z$ coordinate of the table plane. 
\revise{To facilitate the exploration in the case of the \gls{ppolag}, \gls{trpolag} and \gls{pcpo} baselines we loosened the table height constraint to be a pair of inequality constraints that covers the range of $\pm2$ cm around the original equality constraint.} 

\paragraph{Task definition.} \revise{In the air hockey setting the task definition vector $T$ is built using the initial observation of the environment. This vector contains the robot joint positions, joint velocities, puck position, puck orientation, and puck velocities. In this task, this task vector allows us to fully identify the desired trajectory, as the goal state is fixed, i.e., drive the puck into the goal area on the other side of the field.}

\begin{table}[b!]
    \centering
    \caption{Definition of the constraints in Air Hockey Hitting task.}
    \begin{tabular}{c|c|c}
    No. & Name & Definition \\
    \hline
    1-7 & Joint positions & $|q| \leq [2.97, 2.09 , 2.97, 2.09 , 2.97, 2.09 , 3.05]$\\
    8-14 & Joint velocities & $|\dot{q}| \leq [1.48, 1.48, 1.75, 1.31, 2.27, 2.36, 2.36]$\\
    15 & Robot's band collision & $x_{ee} > x_{ab} + r_m$ \\
    16 & Left band collision & $y_{ee} < y_{lb} - r_m$ \\
    17 & Right band collision & $y_{ee} > y_{rb} + r_m$\\
    18 & Table height & $z_{ee} = z_t$
    \end{tabular}
    \label{tab:ah_constraints}
\end{table}

\subsection{Air Hockey real robot deployment}
As in the simulated setup, the real Air Hockey environment is composed of a Kuka Iiwa 14 robot arm (see Figure~\ref{fig:air_hockey_env}). The robot is equipped with an end effector composed of a metal rod, a gas spring, a passive universal joint, and a mallet. The mallet is composed of a movable attachment flange and the mallet itself. The flange is supported by a foam core, allowing for an additional compression of the end effector. This compression is particularly useful to avoid damage in case of small constraint violations and allows the robot to compress slightly the mallet on the table, reducing the probability of flipping. 
Furthermore, we have another Mitsubishi PA10 robot equipped with a suction cup, that is used to reset the puck position on the table.

The Kuka robot is controlled by an Active Rejection Disturbance Controller with a linear trajectory interpolator. The trajectory controller takes the desired position, velocity, and acceleration as input. The interpolator interpolates them linearly into a 1000 Hz command while the learned agent generates the command at 50 Hz. While this interpolation scheme does not produce realizable trajectories, the control frequency of 1000 Hz provides a smoothened command and avoids spikes in the interpolation, which are often problematic for high-order interpolation. A small safety layer is applied to the system to ensure that no wrong commands are applied to the system and that no command is skipped, causing a sudden stop of the commanded trajectory. This is to prevent damage to the robot, but cannot impose any of the safety constraints considered in this paper.

We use the Optitrack motion tracking system to track the puck at 120Hz. 
Different from the simulated environment, we block the goal areas to allow the PA10 arm to easily and automatically reset the puck after each hit attempt.
We reset the puck in a predefined grid of positions. However, the airflow of the air hockey table causes the puck to drift randomly. This makes it impossible to evaluate exactly specific hitting positions with the automatic reset setup.



\section{Samples transformation function}
\label{app:samples_transformation}
To maintain the common scale of the standard deviations computed with $\func_\theta^\sigma(\task)$ and to put bounds on the sampled values, we introduced a transformation $\rho(\zeta)$ of samples $\zeta$.

Let's observe, that we may have several different ways to interpret the the \gls{mp} weights. In the most common setting, they are just the weights of the configuration \glspl{mp}. However, they may also be used to parameterize the time scaling factor $T_s$, control points of the time B-spline $r(s)$ or some boundary parameters, like desired position $\q_d$, velocity $\dq_d$ or acceleration $\ddq_d$. In this case, we may not want to have the same level of exploration noise for all of them. Moreover, we may impose some bounds on the predicted values, such that for example sampled velocity does not exceed the maximal one.
To achieve this, we formulate the $\rho$ function for time-related samples by
\begin{equation}
    \rho_t(\zeta) = \exp(\scale \zeta),
\end{equation}
where $\scale$ is a sample scaling factor.
In the case of the weights related to configuration and its derivatives
we transform the samples with
\begin{equation}
    \rho_q(\zeta) = \Xi \tanh(\scale \zeta),
\end{equation}
where $\Xi$ is the desired bound put on the sampled values.

In particular, in our experiments we have the following scaling factors: (i) for weights related to time $\scale_t = 1$, (ii) for configuration weights $\scale_q = 0.02$, (iii) for end configuration $\scale_{q_d} = 0.02$ in case of no prior knowledge, and $\scale_{q_d} = 0.007$ with prior knowledge, while $\scale_{\dot{q}_d} = 0.02$ and $\scale_{\ddot{q}_d} = 1$.
\revise{Values of these scaling factors were chosen heuristically to achieve reasonable levels of initial exploration.}
In turn, the bounds $\Xi$ for the configuration-related weights are set to $\pi$ rad for the air hockey hitting task, and to $2\pi$ for the heavy object task. When weights are used to parameterize the desired velocity adjustment, then we use $2\dq_{max}$ to allow for completely reversing the velocity bias, while in case of desired accelerations, we set them to $\ddq_{max}$. 

\section{Imposing prior knowledge on the trajectory representation}
\label{app:prior_knowledge}
One of the big benefits of using \glspl{mp} for the trajectory representation is the possibility to impose boundary constraints. In the most general formulation, we can impose the knowledge about the initial configuration on any type of \gls{mp}. Let's note that if we have the trajectory defined by a \gls{mp}, we can determine the first element of the weight vector $\cps_1$, by solving a simple equation
\begin{equation}
    \scalemath{0.9}{\cps_1 = \frac{\BasisFunctions_{:,2:}(0)\cps_{2:}}{\BasisFunctions_1(0)}},
\end{equation}
where index $2:$ means that we skip the first element.
In the case of the basis functions with support equal to $[0; 1]$, like \gls{promp} this is theoretically all we can do. However, in practice, if the first basis function is very close to $0$ for $s=1$, then we can similarly compute the last weight, making only a very small error. To have this error equal to $0$ one needs to have the support of at least one basis function to not contain $0$, like in the \gls{prodmp} case.
However, the most comfortable situation from the boundary conditions point of view is when the supports of the basis functions cover only some overlapping proper subsets of the domain, like for B-splines. In that case, we can fairly easily identify the subsequent boundary weights based on the boundary configurations and their derivatives. Then, if we can find the values of the next derivatives of the phase variable w.r.t. time, we can compute boundary configuration \gls{mp} weights. Thus, we can impose the boundary conditions not only on the configuration but also on velocities, accelerations, and higher-order derivatives. For more details about computing them in the case of the B-spline trajectory representation, we refer the reader to~\cite{kicki2023fast}.

In the next points we discuss what kind of prior knowledge about the task can be applied to different \glspl{mp}, illustrating the usefulness of this feature and explaining in detail the structure of the models used for comparison in Figures~\ref{fig:learning_curves},~\ref{fig:air_hockey_sim} and \ref{fig:kino_stats}.

\subsection{Heavy Object Manipulation}
Heavy object manipulation is an example of a task in which one of the main goals is to reach a certain pose of the manipulated object and maintain it till the end of the episode. In this type of task, typically many possible robot configurations satisfy the task objective. However, one may achieve significantly better results when biasing the end configuration with the one computed with inverse kinematics. We show this phenomenon in Figure~\ref{fig:learning_curves} and \ref{fig:kino_stats}, where all methods that utilize the prior knowledge outperformed their uninformed versions. In the considered task, the goal is not only to reach the target pose but also to stop the robot at that point. This kind of requirement cannot be directly imposed by both \gls{promp} and \gls{prodmp}. However, the proposed B-spline \gls{mp} allows one to set three last configuration \gls{mp} weights in such a way that the last point of the trajectory reaches the goal with zero velocity and acceleration, which ensures smooth stopping.

\subsection{Air Hockey Hitting}
In the air hockey hitting task, imposing the boundary conditions plays also a very important role. While in this case, it is not so obvious how to choose the desired final acceleration, we can provide a good initialization by setting a good hitting configuration and velocity. Using this technique, we can achieve significantly shorter training and obtain the biggest rewards. \revise{In the experiments performed, we did this for the proposed \gls{ours} method, by biasing the end velocity and the hitting configuration} with the values obtained with the optimization procedure proposed in~\cite{liu2021efficient}. Moreover, to compensate for the moving puck, we first computed the time B-spline to know the trajectory duration, and only then, using this duration for predicting the puck pose, we imposed the adjusted boundary conditions achieving decent hitting abilities from the very first episode. It is worth noting that incorporating the velocity level boundary conditions is not so straightforward in the case of \gls{promp} and \gls{prodmp}. Therefore, the use of the B-spline-based \glspl{mp} seems to be a better choice for tasks that require dynamic nonprehensile manipulation, like hitting the puck or tossing objects.

It is worth noting that the very important part of the air hockey hitting task, especially in terms of constraint satisfaction, is the stopping phase. After hitting the puck with a very high velocity robot needs to slow down its motion and at the same time still satisfy the table plane constraint, while potentially being in the state space areas of reduced manipulability. To handle this phenomenon, we leveraged the possibility of smoothly composing multiple trajectories. Instead of generating only the hitting motion we generate at the same time both hitting and stopping motion, which connects near to the hitting point with continuous accelerations.

\subsection{\revise{Choice of scaling factors}}
\revise{One of the key parameters of the black box policy optimization algorithm is the sample scaling factors, that take into account the desired level of the initial exploration.} 

\revise{While in general, this parameter can be set arbitrarily, in robotics settings these parameters are quite easy to tune.
Indeed, the initial exploration in the task space should match the scale of the used robots and considered tasks. Therefore, given the initial values of the variance of the normal distribution generated by the neural network, we decided on the scaling factor for the MP weights. In the experiments for this paper, the desired end effector variance is in the order of decimeters.}

\revise{However, some of the elements of the action space may have special meanings, like end configuration, velocity, or end acceleration. For these parameters, we may want to assign them different scaling factors keeping in mind that they represent different quantities, such as velocity and acceleration. These quantities may require different levels of exploration. 
For this reason, in our experiments, we decided to set a relatively high value for the acceleration--- as we are very uncertain about its initial bias and the scale of the accelerations is much bigger than configurations ---and a relatively low value for velocities--- as we believe that the desired end velocity is very accurately given by the knowledge about the task.
Similar reasoning was conducted also for the scaling factors associated with the time-scaling B-spline control points, but in this case, we focused on the desired level of exploration in terms of trajectory duration, which we wanted to be about 10\% of the mean initial trajectory duration.}

\revise{Thanks to the used parametrization, especially the B-spline-based one, the process of setting these parameters is quite intuitive due to the physical meaning of the considered quantities. Moreover, one can easily control the level of exploration by simulating a batch of random trajectories with the considered scaling factors and visually observing if the initial exploration matches the expectations.}

\section{\revise{Motion primitives flexibility}}
\revise{
One of the goals of this paper was to bridge the gap between the \gls{saferl} and \glspl{mp}. Therefore, it is worth considering whether and to what extent we limited the generality of the classical \gls{cmdp} framework by establishing a coupling with \glspl{mp}.}

\revise{
We argue that the use of motion primitives does not introduce excessive limitations into the \gls{cmdp} framework but rather limits the space of learnable policies and allows for an easy introduction of inductive biases. Nevertheless, there is an important difference w.r.t. classical \gls{cmdp} methods, which is the trajectory-level (\glspl{mp} requires black box learning) vs step-based exploration (classical \gls{cmdp}).
However, in general, there is no difference in terms of behavior we can achieve with black-box optimization (excluding the above-mentioned inductive biases) and classical \gls{cmdp} formulation. The key difference is that step-based exploration exploits local information (possibly allowing faster learning) and black box exploration allows us to explore better, which is particularly useful in settings with sparse reward and complex constraints as the tasks presented in this paper, where classical \gls{cmdp} methods are struggling.
Finally, it is also possible to learn with \glspl{mp} with classical \gls{cmdp} methods, by adding Gaussian noise at the trajectory level, but this comes at the cost of the smoothness of the trajectory.}

\revise{
The main limitation that \glspl{mp}, in general, introduces is the reduction in the space of learnable policies. Fortunately, this seems to not be very restrictive as in the literature we can see a broad range of applications in which \glspl{mp} provided sufficient flexibility~\cite{muelling2010learning,maeda2014learning,parisi2015reinforcement, ploeger2021high,celik2022specializing,otto2023deep,li2023prodmp,drolet2023learning}. In fact, the flexibility of every \gls{mp} can be controlled by the number of basis functions used. However, the shape and distribution of the basis functions along the phase variable axis may affect the expressiveness of the particular \gls{mp}.
For example, \gls{prodmp} seems to be very effective in tasks where the final pose of the end-effector is particularly important, given the distribution of the basis functions and their unbalanced scale. 
Instead, the B-spline-based representation is characterized by flexibility along the entire trajectory and shines when we have complex constraints to impose due to the decoupling of the geometric path and the temporal path.
}

\section{Experimental details}
While the general description of the performed experiments is included in the main text, we provide the details about them here to increase the reproducibility of our research.

First, as we mentioned in Section~\ref{sec:constrained_rl}, \gls{ours} is meant to generate the trajectories, so it requires a controller to generate actions. In this paper, we used the proportional-derivative controller with feed-forward implemented in the Air Hockey Challenge repository~\cite{ahc} with default gains. The same controller is used by default by the \gls{atacom} baseline. In turn, for the \gls{ppolag}, \gls{trpolag} and \gls{pcpo} we used the inverse dynamics algorithm from MuJoCo to transform the accelerations predicted by the policy into the torques.

In Table~\ref{tab:architectures}, we present the architectures of all the neural networks used in the experiments, except the one used by \gls{ours}, as it is slightly more complex than a sequence of \gls{fc} layers. In this case, the input is processed first by 3 \gls{fc} layers, and then the resultant representation is used by two heads: (i) configuration head with 2 \gls{fc} layers, and (ii) time head with a single layer. Each layer, except the output ones, consists of 256 neurons and a $\tanh$ activation function. 

\begin{table}[b!]
    \centering
    \caption{Neural network architectures.}
    \begin{tabular}{c|cc}
    Method & Hidden layers & Activation\\
    \hline
    Value network of \gls{ours} (all variants) & $4 \times 256$ &  $\tanh$ \\
    Policy network of \gls{ours}-\gls{promp}/\gls{prodmp} & $4 \times 256$ & $\tanh$ \\
    Actor network of \gls{atacom} & $3 \times 128$ & $\selu$\\
    Critic network of \gls{atacom} & $3 \times 128$ & $\selu$\\
    Actor network of \gls{ppolag} & $2 \times 256$ &  \revise{$\tanh$}\\
    Critic network of \gls{ppolag} & $2 \times 256$ &  \revise{$\tanh$}\\
     \revise{Actor network of \gls{trpolag}} &  \revise{$2 \times 256$} &   \revise{$\tanh$}\\
     \revise{Critic network of \gls{trpolag}} &  \revise{$2 \times 256$} &  \revise{$\tanh$}\\
     \revise{Actor network of \gls{pcpo}} &  \revise{$2 \times 256$} &   \revise{$\tanh$}\\
     \revise{Critic network of \gls{pcpo}} &  \revise{$2 \times 256$} &  \revise{$\tanh$}\\
    \end{tabular}
    \label{tab:architectures}
\end{table}

Many of the chosen learning hyperparameters are common for both heavy object and air hockey hitting tasks, as well as for the considered methods. We list them in Table~\ref{tab:common_hyperparams}. 
However, some of the parameters are specific to the given algorithm, thus, we list them in Tables~\ref{tab:ours_hyperparams},~\ref{tab:atacom_hyperparams}, and \ref{tab:ppolag_hyperparams}.

\begin{table}[ht]
    \centering
    \caption{Common hyperparameters for all experiments}
    \begin{tabular}{c|c}
    Hyperparameter & Value \\ \hline
    Number of episodes per epoch &  \revise{$64$}\\
    \revise{ Number of fits per epoch} &  \revise{$32$}\\
     \revise{Number of batches per fit} &  \revise{1}\\
     \revise{Batch size} &  \revise{$64$}\\
    Number of evaluation episodes & $25$\\
     \revise{$\gamma$} &  \revise{$0.99$}\\
    $\varepsilon_\text{PPO}$ & $0.05$\\
    \end{tabular}
    \label{tab:common_hyperparams}
\end{table}

\begin{table}[ht]
    \centering
    \caption{\gls{ours} hyperparameters}
    \begin{tabular}{c|c}
    Hyperparameter & Value \\ \hline
    Number of configuration weights & $11$\\
    Number of time weights & $10$\\
    Initial standard deviation & $1$\\
    Constraint learning rate $\alpha$ & $0.01$\\
    Manifold metric decline bound $\beta$ & 0.1\\
    Policy learning rate (\gls{ours} all variants) & $5\cdot10^{-5}$\\
    Value function approximator learning rate & $5\cdot10^{-4}$\\
    $\alpha$ & \\
    \end{tabular}
    \label{tab:ours_hyperparams}
\end{table}

\begin{table}[ht]
    \centering
    \caption{\gls{ppolag},  \revise{\gls{trpolag} and \gls{pcpo}} hyperparameters}
    \begin{tabular}{c|c}
    Hyperparameter & Value \\ \hline
    Actor learning rate & $5\cdot10^{-4}$\\
    Critic learning rate & $5\cdot10^{-4}$\\
    Lagrangian multiplier learning rate & $0.01$\\
    Cost limit (heavy object) & 10\\
    Cost limit (air hockey) & 0.01\\
     \revise{$\lambda_{GAE}$} &  \revise{0.95}\\
    \end{tabular}
    \label{tab:ppolag_hyperparams}
\end{table}

\begin{table}[ht]
    \centering
    \caption{\gls{atacom}+\gls{sac} hyperparameters}
    \begin{tabular}{c|c}
    Hyperparameter & Value \\ \hline
    Actor learning rate & $3\cdot10^{-4}$\\
    Critic learning rate & $3\cdot10^{-4}$\\
    Replay buffer size & $2\cdot10^5$\\
    Soft updates coefficient & $0.001$\\
    Warmup transitions & $10^4$\\
    Learning rate of $\alpha_{SAC}$& $5\cdot10^{-5}$\\
    Target entropy & -2 \\
    \end{tabular}
    \label{tab:atacom_hyperparams}
\end{table}



\clearpage
\section{Additional results on the heavy object task}
\begin{figure}[hb]
    \centering \includegraphics[width=\textwidth]{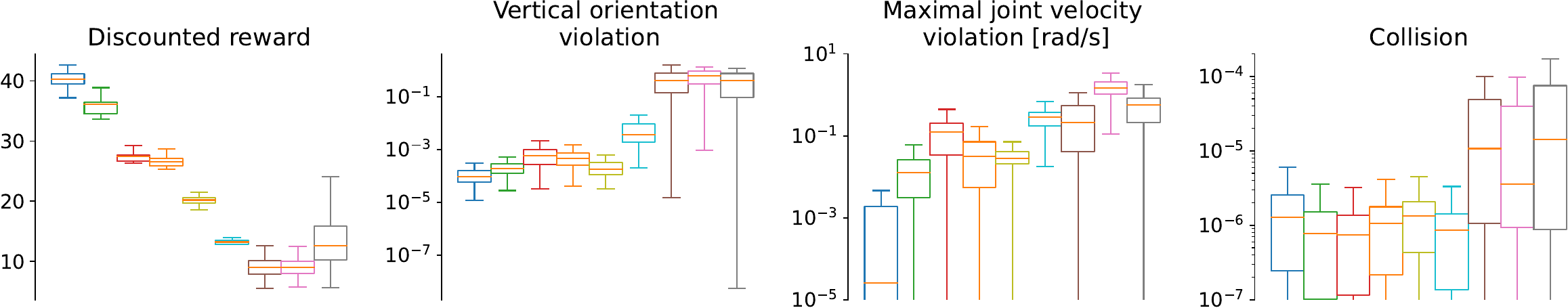}
    \includegraphics[width=\textwidth]{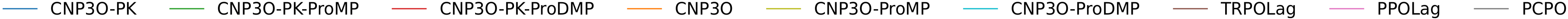}
    \caption{Statistical analysis of the considered approaches on the heavy object task}
    \label{fig:kino_stats}
\end{figure}

In this section, we present further analysis of the heavy object task, by looking at the metrics at the end of learning. We present in Figure~\ref{fig:kino_stats} the boxplot showing the distribution of metrics of 100 episodes for each seed.
Our results show that, in the setting with prior knowledge, the best parameterization for the motion primitives is the B-splines, achieving better performance and, in general, lower constraint violations than any other method. 
Adding prior knowledge is always an advantage for learning performance on all metrics, with the notable exception of B-Splines for collision penalty. However, it is worth noting that the collision penalty is extremely low compared to the other metrics, and most approaches can avoid collision robustly.
Without prior knowledge, the B-spline parametrization obtains slightly worse constraint violations but also achieves slightly better performance. However, we remark that the B-splines can learn faster than the other motion primitives in this setting.
\vspace{2cm}

\end{document}